\documentclass[journal]{IEEEtran}
\usepackage{graphicx} 
\usepackage{epstopdf}
\usepackage{amsmath,amsfonts}
\usepackage{algorithmic}
\usepackage{amsthm}
\usepackage{algorithm}
\usepackage{array}
\usepackage[dvipsnames, svgnames, x11names,table,xcdraw]{xcolor}
\usepackage{multirow}
\usepackage{comment}
\usepackage{color}
\usepackage{textcomp}
 \usepackage[table,xcdraw]{xcolor}
\usepackage{stfloats}
\usepackage{url}
\usepackage{verbatim}
\usepackage{makecell}
\usepackage{graphicx}
\usepackage{booktabs}
\usepackage[table,xcdraw]{xcolor}
\usepackage[normalem]{ulem}
\useunder{\uline}{\ul}{}
\usepackage{graphicx}
\usepackage{amsmath, bm} 
\usepackage{multirow}
\usepackage{adjustbox}
\usepackage{subcaption}
\usepackage{stfloats} 
\usepackage{float}
\usepackage{hyperref}
\usepackage[table]{xcolor}
\definecolor{bg01}{RGB}{240,240,255} 
\definecolor{bg02}{RGB}{255,240,240} 
\definecolor{bg03}{RGB}{255,255,230} 
\usepackage{cite}

\newtheorem{theorem}{Theorem}

\def\x{\mathbf{x}}

\begin{document}
\setlength{\belowdisplayskip}{4pt}
\title{COLI: A Hierarchical Efficient Compressor for Large Images}

\author{Haoran Wang$^{1}$, Hanyu Pei$^{1}$, Yang Lyu$^{2}$, Kai Zhang$^{1}$, Li Li$^{3}$, Feng-Lei Fan$^{1*}$
\thanks{This work did not involve human subjects or animals in its research.}
\thanks{Corresponding author: Feng-Lei Fan (fenglfan@cityu.edu.hk and hitfanfenglei@gmail.com)}
\thanks{$^{1}$Haoran Wang, Hanyu Pei, Kai Zhang (kaizhang0116@163.com), and Feng-Lei Fan are with Frontier of Artificial Networks (FAN) Lab, Department of Data Science,
City University of Hong Kong, Hong Kong, China SAR}
\thanks{$^{2}$Yang Lv (yang.lv@united-imaging.com), Molecular Imaging Business Unit, Shanghai United Imaging Healthcare Co., Ltd, Shanghai, China}
\thanks{$^{3}$Li Li is with the MoE Key Laboratory of Brain-Inspired Intelligent Perception and Cognition, University of Science and Technology of China, Hefei 230027, China.}

}

\markboth{submitted to IEEE Transactions on Radiation and Plasma Medical Sciences, Vol. XX, No. XX, XX 2024}%
{Shell \MakeLowercase{\textit{et al.}}: A Sample Article Using IEEEtran.cls for IEEE Journals}


\maketitle

\begin{abstract}

The escalating adoption of high-resolution, large-field-of-view imagery amplifies the need for efficient compression methodologies. Conventional techniques frequently fail to preserve critical image details, while data-driven approaches exhibit limited generalizability. Implicit Neural Representations (INRs) present a promising alternative by learning continuous mappings from spatial coordinates to pixel intensities for individual images, thereby storing network weights rather than raw pixels and avoiding the generalization problem. However, INR-based compression of large images faces challenges including slow compression speed and suboptimal compression ratios. To address these limitations, we introduce COLI (Compressor for Large Images), a novel framework leveraging Neural Representations for Videos (NeRV). First, recognizing that INR-based compression constitutes a training process, we accelerate its convergence through a pretraining-finetuning paradigm, mixed-precision training, and reformulation of the sequential loss into a parallelizable objective. Second, capitalizing on INRs' transformation of image storage constraints into weight storage, we implement Hyper-Compression—a novel post-training technique—to substantially enhance compression ratios while maintaining minimal output distortion. Evaluations across two medical imaging datasets demonstrate that COLI consistently achieves competitive or superior PSNR and SSIM metrics at significantly reduced bits per pixel (bpp), while accelerating NeRV training by up to 4×.

\end{abstract}


\IEEEpeerreviewmaketitle

\section{Introduction}

\IEEEPARstart{W}{ith} the explosion of ultra-high-resolution and large-field-of-view images and videos, the demand for efficient storage and transmission methods for large images has intensified.~\cite{toderici2017full} In contemporary applications such as medical imaging, the high-resolution 3D imagery plays a pivotal role. However, such images often come with an extremely large file size that poses considerable challenges in terms of storage, bandwidth, and processing efficiency.~\cite{20123D} A good image compression algorithm can significantly alleviate storage requirements and lower transmission costs. Conventional image compression algorithms, including JPEG\cite{wallace1992jpeg}, JPEG2000~\cite{taubman2002jpeg2000}, and WebP~\cite{googlewebp}, rely on hand-engineered transformations, quantization, and entropy coding schemes. Although these methods are well-established and computationally efficient, they generally struggle to preserve fine image details at high compression ratios. Their fixed transformation pipelines are also limited in capturing complex semantic structures within images. Furthermore, entropy coders such as Huffman coding \cite{huffman1952method}, while effective in reducing symbol redundancy, inherently rely on discrete symbol modeling and lack mechanisms for capturing spatial and contextual dependencies across images.


Traditional image compression algorithms, while mature and computationally efficient, often struggle to preserve fine-grained details and adapt to diverse image characteristics, especially under high compression demands. To address these limitations, learning-based approaches have been introduced, leveraging data-driven priors to better capture complex spatial and semantic structures. With the rise of deep learning, the field of image compression has undergone a paradigm shift. Neural network-based methods, \textit{i.e.}, convolutional autoencoders \cite{balle2018variational}, generative models \cite{mentzer2020high}, and attention-based architectures, enable the learning of compact hierarchical representations directly from data. These data-driven approaches have demonstrated impressive gains in perceptual quality and coding efficiency. However, they come with intrinsic limitations: these models often exhibit poor generalization to unseen resolutions or domains, and typically suffer from scalability issues when applied to extremely large images. For the latter issue, many convolutional neural network (CNN)-based solutions rely on patch-wise encoding or full-image processing with fixed-size architectures, both of which limit their practicality for high-resolution inputs such as $12,000\times12,000$ images.

In response to these challenges, a novel line of research has explored the use of Implicit Neural Representations (INRs), which encode images or videos as continuous functions rather than discrete pixels. These models learn a neural mapping from spatial coordinates to RGB values, expressed as \( [f_\theta^r(x, y),f_\theta^g(x, y), f_\theta^b(x, y)] \rightarrow [p_r,p_g,p_b] \), where the network parameters encode the signal \cite{sitzmann2020implicit, tancik2020fourier}. This allows for resolution-agnostic storage and continuous reconstruction, with no generalization issue. Prominent INR models, such as SIREN \cite{sitzmann2020implicit}, NeRF \cite{mildenhall2020nerf}, and Neural Representations for Videos (NeRV) \cite{chen2021nerv} have demonstrated strong representational capabilities with a compact form, which hold great potential as compressors. Recent NeRV-style codecs have been widely used for frame-wise representation and compression in videos, mapping frame indices to full frames as a standard INR paradigm \cite{zhao2025treenerv}. Though originally designed for videos, NeRV’s architectural design makes it a compelling option for static image compression. In addition to these architectures, recent INR-based compression methods such as the Coordinate-based Image Network (COIN) \cite{dupont2021coin} and its extension COIN++ \cite{dupont2022coinpp} were also proposed for static images. COIN fits a lightweight coordinate MLP to an image and stores its quantized parameters as the compressed representation, while COIN++ further introduces learned low-bit modulations to reduce storage cost. These approaches essentially leverage model compression techniques to improve the quality of image compression, with the INR as a bridge, which becomes a common practice in INR-based compression.

\textbf{However, applying INR models to compress large images still faces severe challenges:} 
\begin{itemize}
    \item \textbf{Slow compression time.} Using an INR network to compress images is a process of training. It requires creating a mapping for every pixel via tuning weights of the INR network, which is computationally expensive. Optimizing such a network on large images further escalates the computational cost, which makes training an INR model for a large image a long process. This drawback limits the scalability of INRs in scenarios where large-scale images need to be compressed in a short time such as hospitals.
    \item \textbf{High compression ratio.} Achieving a high compression ratio without significant quality degradation presents a fundamental challenge for INR model, particularly for large images. This stems from the inherent representational characteristics of typical INR models, which prioritize smooth signal reconstruction and struggle to efficiently encode high-frequency components. Consequently, pushing the compression ratio beyond moderate levels leads to the loss of relevant features or blurring textures. This limitation directly hinders the practical employment of INR on large images.
\end{itemize}

\begin{figure*}[htbp]
    \centering   \includegraphics[width=0.95\linewidth]{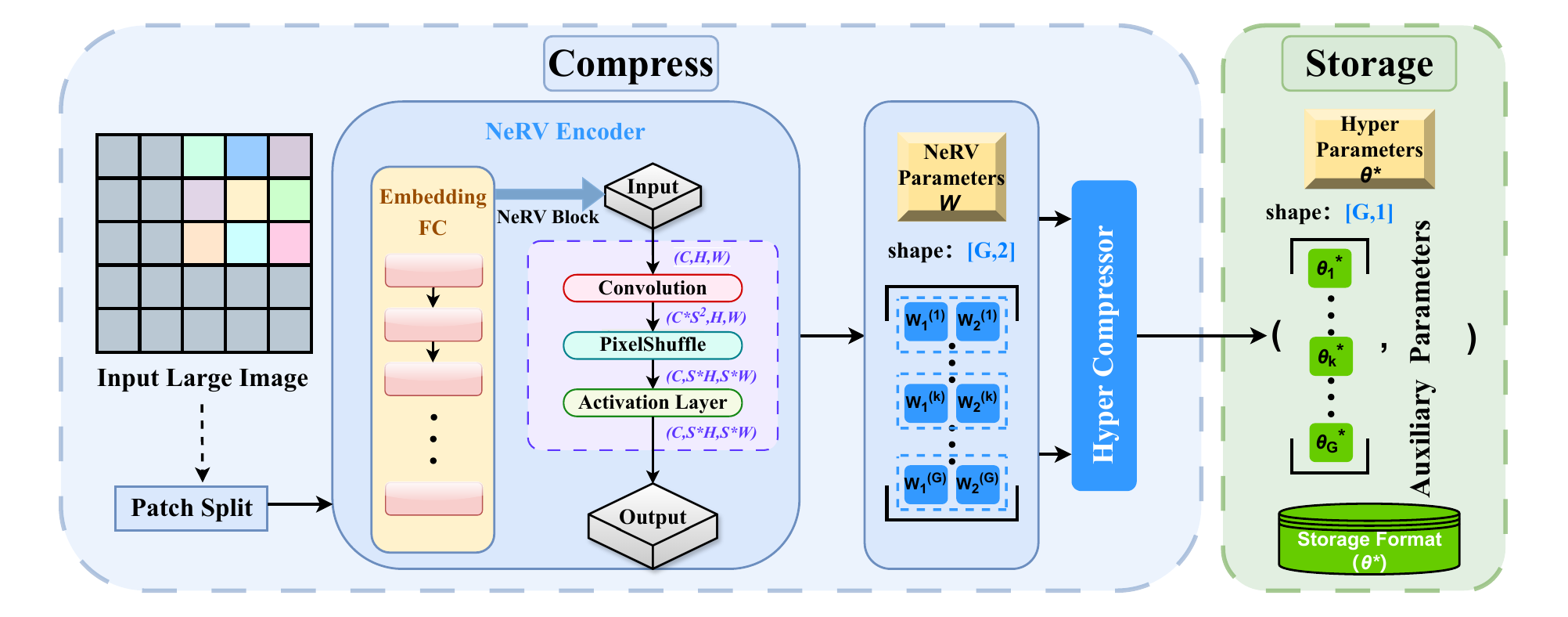}
        \caption{The schematic diagram of the COLI framework. NeRV encodes a large image into weights, which are flattened and reshaped into a matrix with the size of \((G,2)\); Hyper-Compression maps each row to one hyperparameter \(\theta^{*}\) for storage and later reconstruction.}
    \label{fig:coli-pipeline}
    \vspace{-0.5cm}
\end{figure*}

In this work, we address the above two issues by proposing a hierarchical efficient \underline{\textbf{CO}}mpressor for \underline{\textbf{L}}arge \underline{\textbf{I}}mages (COLI) based on NeRV that systematically tackles both computational and compression efficiency challenges. 

\textbf{To address the computational cost challenge}, we introduce a pretraining-finetuning paradigm at the beginning of COLI that enables the INR network to learn shared features before adapting to specific images, significantly reducing convergence epochs. Subsequently, we implement mixed precision training coupled with decoupled patch-based loss computation to enable GPU parallelism and reduce per-epoch training time in training NeRV. Furthermore, we achieve substantial throughput improvements by parallelizing multiple INR network training processes within the same GPU architecture. This comprehensive acceleration framework transforms the computational bottleneck from a sequential training constraint into a parallelizable optimization task.

\textbf{To further enhance compression efficiency}, we integrate a dedicated Hyper-Compression module \cite{fan2024hyper} that applies post-training model compression techniques specifically tailored for INR networks. This approach employs a hyperfunction $h(\theta; n)$ to map high-dimensional parameter vectors $\{w_n\}_{n=1}^N$ to compact representations controlled by a single hyperparameter $\theta$, ensuring that $|w_n - h(\theta^*; n)| < \epsilon$ for an arbitrarily small $\epsilon$. The key advantage lies in that $\theta^*$ requires significantly less storage than the original parameter set $\{w_n\}_{n=1}^N$, and can be further optimized by setting $\theta^*$ to integer values, substantially reducing storage requirements. Unlike conventional compression techniques such as pruning or quantization, Hyper-Compression achieves superior compression ratios without degrading reconstruction quality while eliminating the need for network retraining or complex optimization procedures.

Figure~\ref{fig:coli-pipeline} provides an overview of the COLI framework. The input image is first divided into patches and encoded by a NeRV-based encoder, which produces neural weights for each patch. These weights are then passed through a hypercompression module, which transforms them into a compact set of hyperparameters \( \theta^* \) for storage. During decompression, the stored \( \theta^* \) is decompressed into the weights of the INR network. These reconstructed weights are then fed into the NeRV decoder, which merges the patches and reconstructs the final image. The entire pipeline forms a compact, end-to-end framework that supports efficient encoding, storage, and decoding of large images.

We systematically evaluate COLI on both cell-level electron microscopy and medical CT datasets, demonstrating its effectiveness for large image compression in terms of both computational efficiency and compression ratio. 
Our multi-facet acceleration strategy achieves up to 8.91$\times$ faster training for comparable PSNR levels (e.g., training time reduced from 98 minutes to 11 minutes at PSNR $\approx$30), while still maintaining around 4$\times$ speedup even at higher PSNR levels with better visual quality.
Meanwhile, the proposed Hyper-Compression module enables significant bit-per-pixel (bpp) reduction while maintaining stable reconstruction quality. 
For example, NeRV combined with Hyper-Compression reduces bpp from 3.99 to 1.06 with minimal visual quality loss (PSNR of 34.37~dB, SSIM of 0.9099). 
Furthermore, COLI achieves lower bpp than NeRV and even conventional codecs, while producing reliable perceptual quality for large biomedical images. In summary, our contributions are threefold:

\begin{itemize}
    \item We present COLI, a novel INR-based large image compression framework that systematically addresses both computational and compression efficiency challenges. By integrating patch-wise INR encoding with a hierarchical Hyper-Compression module, COLI reduces redundant network parameters without the need for retraining.
    \item We design an effective multi-facet acceleration strategy that achieves significant training speedup through pretraining-finetuning, mixed-precision training, and batch-parallel scheduling. This makes INR-based image compression practical for real-world large-scale datasets.
    \item We conduct comprehensive experiments on diverse high-resolution datasets. Results consistently verify that COLI delivers competitive or better PSNR and SSIM at substantially lower bpp, demonstrating strong applicability for storing and transmitting large biomedical images.
\end{itemize}

\section{Related Work}

\subsection{Methods for Image Compression} 

Traditional image compression methods such as JPEG\cite{wallace1992jpeg}, JPEG2000~\cite{taubman2002jpeg2000}, and WebP~\cite{googlewebp} have dominated the field for decades due to their computational efficiency and widespread compatibility. However, these methods face fundamental limitations when applied to large-field-of-view, high-resolution images. Traditional codecs rely on fixed-size block processing, leading to visible blocking artifacts that become increasingly problematic at high compression ratios—particularly critical for applications requiring fine detail preservation such as medical imaging. Moreover, their fixed transformation pipelines (DCT, DWT) cannot adapt to diverse content characteristics, limiting compression efficiency on complex scenes or domain-specific imagery.

Learning-based compression methods have emerged to address these limitations through end-to-end optimization. Autoencoder-based approaches \cite{balle2018variational}, GANs \cite{mentzer2020high}, and transformer architectures \cite{cheng2020learned,zou2022devil,zhu2022transformer,liu2023learned} have achieved superior rate-distortion performance by learning adaptive representations. However, these methods predominantly rely on autoregressive priors for accurate entropy modeling, which necessitates serial decoding and leads to prohibitive latency when processing large images. Furthermore, these approaches suffer from generalization issues—their performance degrades significantly on images that deviate from training distributions, limiting their practical deployment in diverse real-world scenarios.

Both traditional and learning-based methods struggle with ultra-high-resolution images due to memory constraints and computational overhead, motivating the need for fundamentally different representation paradigms that can efficiently handle large-scale visual content.

\subsection{Implicit Neural Representation}

INRs offer a paradigm shift by encoding images as continuous functions parameterized by neural networks. Instead of storing pixel data or compressed features, INRs store only network parameters that define a mapping from spatial coordinates $(x, y)$ to signal values, enabling resolution-agnostic and potentially highly compact representations.

Early INR approaches established the foundation for this paradigm but suffered from computational limitations. SIREN \cite{sitzmann2020implicit} introduced periodic sine activation functions to enhance the network's ability to capture high-frequency details, but its fully-connected architecture leads to prohibitive computational costs when processing high-resolution images that require millions of coordinate queries. Similarly, NeRF \cite{mildenhall2020nerf} demonstrated the power of coordinate-based MLPs in learning complex continuous functions for 3D scene modeling, though it suffers from long training time and inference latency. BACON \cite{lindell2021bacon} addressed multi-scale representation through hierarchical processing with band-limited filters, enabling a better capture of both global and local image structures, yet it still relies on computationally expensive coordinate-wise processing.

There were INR methods developed specifically for compression, focusing not on altering the coordinate network itself but on improving the storage efficiency of its parameters through quantization. COIN \cite{dupont2021coin} fits a lightweight SIREN-style MLP to each image and applying low-bit quantization to its weights, thereby reducing the parameter storage cost. COIN++ \cite{dupont2022coinpp} further introduces a meta-learned base network along with learned low-bit modulation parameters, which are more amenable to quantization and entropy coding than full network weights. Collectively, these methods enhance the compression capability of INR models by employing more efficient quantization schemes, shedding light on combining model compression techniqus to improve INR-based codecs.

A significant progress was made by NeRV \cite{chen2021nerv}, which eliminated the computational bottleneck of pixel-wise coordinate regression by adopting a convolutional decoder architecture. Rather than querying coordinates individually, NeRV generates images from implicit index codes via lightweight decoders, achieving fast inference while avoiding coordinate-wise processing overhead. Building upon NeRV's foundation, MetaNeRV \cite{chen2022metanerv} incorporates meta-learning frameworks to improve adaptability across diverse content, while Implicit Neural Teaching (INT) \cite{zhang2023int} achieves over 30\% training time reduction through selective sampling of informative signal fragments.

Despite these advances, applying INR models to compress large images faces critical challenges. The computational cost becomes prohibitive for ultra-high-resolution images, and achieving high compression ratios without quality degradation remains challenging due to INRs' inherent bias toward smooth signal reconstruction and difficulty in efficiently encoding high-frequency components. These limitations motivate our choice of NeRV as the foundation for COLI. By combining NeRV's efficient architecture with hypernetwork-based parameter compression, COLI can achieve the aggressive compression ratios required for practical large-image compression while maintaining reconstruction quality and computational tractability. 

\subsection{Post-training Model Compression}

Post-training compression techniques are essential for deploying a network efficiently, with pruning~\cite{han2015learning}, quantization~\cite{jacob2018}, and low-rank decomposition~\cite{denton2014exploiting}, being prominent. Pruning pursues the sparsity by eliminating redundant weights (unstructured) or entire neurons/channels (structured) from a trained model to reduce the model size. It can accelerate inference and lower memory footprint, but involves potential accuracy loss, irregular sparsity requiring specialized hardware for speedups, and iterative fine-tuning complexity. Quantization compresses models by reducing the numerical precision of the weights / activations (e.g., 32-bit floats $\to$ 8-bit integers), offering both memory savings and hardware-friendly speedups on compatible processors. However, it is hard to achieve low-bit quantization. Low-rank decomposition factorizes a dense weight matrices (e.g., via SVD) into a product of singular matrices, reducing parameters and operations for faster computation. While effective for layers with high-rank redundancy, it is subjected to significant accuracy drops if ranks are overly reduced, and decomposition complexity. 

In this work, we introduce a novel post-training compression method referred to as the Hyper-Compression~\cite{fan2024hyper} method for the INR network. The Hyper-Compression has a reasonably good compression ratio and can compress a network in a fast manner. These merits make it occupy a niche for further compressing the INR network resulting from a large image, which particularly needs higher compression ratio and fast compression time.

\section{Compressor for Large Image(COLI)}

In this work, we propose a novel compression method, referred to as COLI, that enjoys two innovations: hierarchical compression and training acceleration.

    \begin{figure}[htbp]
        \centering
        \includegraphics[width=0.5\textwidth]{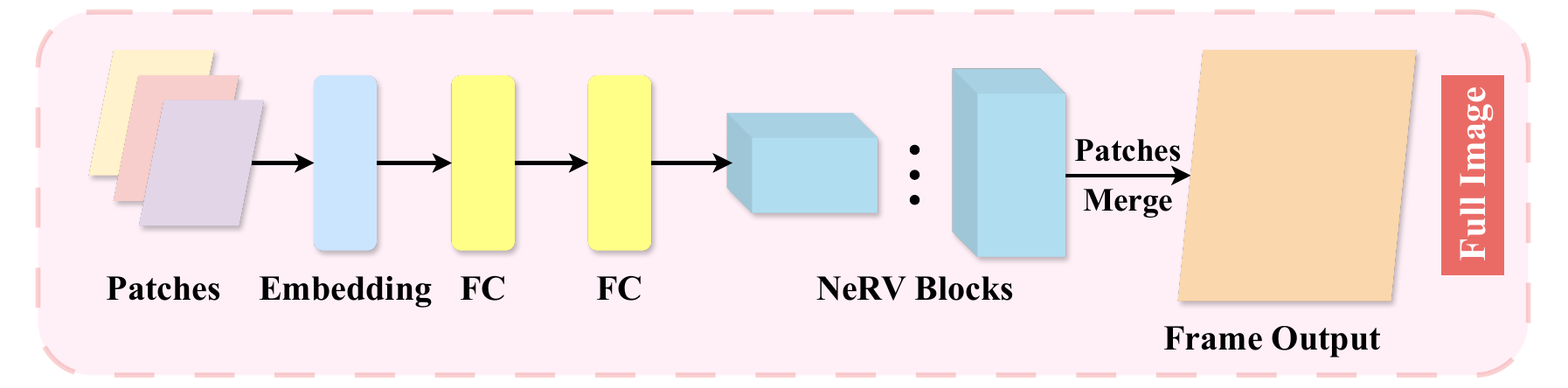} 
        \caption{The overall structure of the NeRV model. The input patches are firstly embedded, then passed through several fully connected layers and a sequence of NeRV blocks, and finally decoded into frame outputs.}
        \label{fig:nerv-structure}
        \vspace{-0.5cm}
    \end{figure}
    
\subsection{Accelerating the Training of NeRV}

Now, we first introduce NeRV and then provide its acceleration scheme. 
    
\subsubsection{\textbf{The NeRV Overview}} 
We begin by dividing the input image \( I \in \mathbb{R}^{H \times W} \), where \( H \) and \( W \) denote the height and width of the image, into non-overlapping rectangular patches along the spatial dimensions. Each patch has a fixed size of \( P_H \times P_W \), where \( P_H \) and \( P_W \) represent the patch height and width, respectively.

Let \( \mathbf{I}_i \in \mathbb{R}^{P_H \times P_W} \) be the \( i \)-th patch, where \( i \in \{1, 2, \dots, N\} \), and the total number of patches \( N \) is computed as:
\begin{equation}
   N = \left\lceil \frac{H}{P_H} \right\rceil \times \left\lceil \frac{W}{P_W} \right\rceil,
\end{equation}

All patches share a common NeRV network for joint representation learning. Specifically, each patch index acts as a conditional input, similar to the frame index in the original NeRV design for videos. This allows the network to learn a continuous mapping that generates each patch’s content while sharing parameters across all patches, thereby achieving global spatial consistency.

Figure~\ref{fig:nerv-structure} illustrates the overall architecture of NeRV. The patch index and positional embedding are first projected into a feature space, then processed by two fully connected layers to expand the feature dimension, followed by several NeRV blocks that progressively refine the latent features, and finally decoded into output patches. Each NeRV block consists of a convolutional layer, a pixel shuffle operation for spatial upsampling, and a non-linear activation function. This design allows NeRV to effectively model high-resolution structures with compact shared weights.

For compression, the NeRV model implicitly encodes the entire image by storing its learned network parameters and the patch indexing logic. This eliminates the need to store pixel data directly. During reconstruction, the decoder uses the same patch index to recover each patch:
\begin{equation}
   \hat{\mathbf{I}}_i = g_{\phi}(f_{\theta}(i)),
\end{equation}
where \( f_{\theta} \) and \( g_{\phi} \) denote the shared encoder and decoder with parameters \( \theta \) and \( \phi \), respectively. The training optimizes the reconstruction loss:
\begin{equation}
   L_{\text{reconstruction}} = \sum_{i=1}^{N} \left\| \mathbf{I}_i - \hat{\mathbf{I}}_i \right\|_2^2,
\end{equation}

Finally, the full image \( \hat{I} \) is reconstructed by stitching all recovered patches together. 
Notably, our approach draws inspiration from NeRV’s original design for video frame representation, which inherently models temporal continuity across adjacent frames. We extend this concept to large static images by dividing them into spatial patches and treating each patch index analogously to a frame index. This perspective effectively transfers the idea of smooth frame transitions in videos to smooth spatial transitions across image regions. Such a continuity-aware representation naturally aligns with biomedical datasets, such as CT or MRI scans, which are typically acquired as sequential slices with strong inter-slice structural coherence. By leveraging NeRV’s sequential continuity principle, our method maintains global spatial consistency across patches, enabling efficient yet coherent compression for high-resolution and medical imaging scenarios.

\subsubsection{\textbf{Reducing Training Time}} Because each NeRV is designed for a single image, scaling the framework to large usage requires faster training and deployment. We propose a three-stage acceleration strategy to improve training efficiency.

a)~\underline{Reducing training epochs via pretrained models}. In practical applications, a batch of images often shares similar structural patterns, \textit{i.e.}, medical images of the same organ share repeated parts or consistent surface textures. Therefore, training NeRV does not need to start from scratch for each new image. We adopt a \textit{pretrained model initialization} strategy. Specifically, a generic NeRV is first trained on a large domain dataset. When compressing a new image, the model parameters $\theta$ are initialized with the pretrained weights $\theta_0$ rather than random initialization. This warm start can reduce the number of training epochs, since initial parameters already represent the common features, and the pre-trained model only needs to adapt the specific image.

The optimization target for NeRV can be generally written as
\begin{equation}
\mathcal{L}_{\text{recon}}(\mathbf{\xi}) = \frac{1}{N} \sum_{i=1}^{N} \big\| g_{\xi}(\mathbf{z}_i) - \mathbf{I}_i \big\|^2,
\end{equation}
Here, $\mathbf{z}_i$ represents the encoded coordinates or latent inputs, $f_{\xi}(\cdot)$ denotes the INR function, and $\mathbf{I}_i$ is the target pixel value. Unlike explicit transfer learning with a regularization term like $\| \xi - \xi_0 \|^2$, our approach uses the pretrained $\xi_0$ purely as a better initialization, without requiring additional constraints.

\begin{figure*}[htbp]
        \centering
        \includegraphics[width=\textwidth]{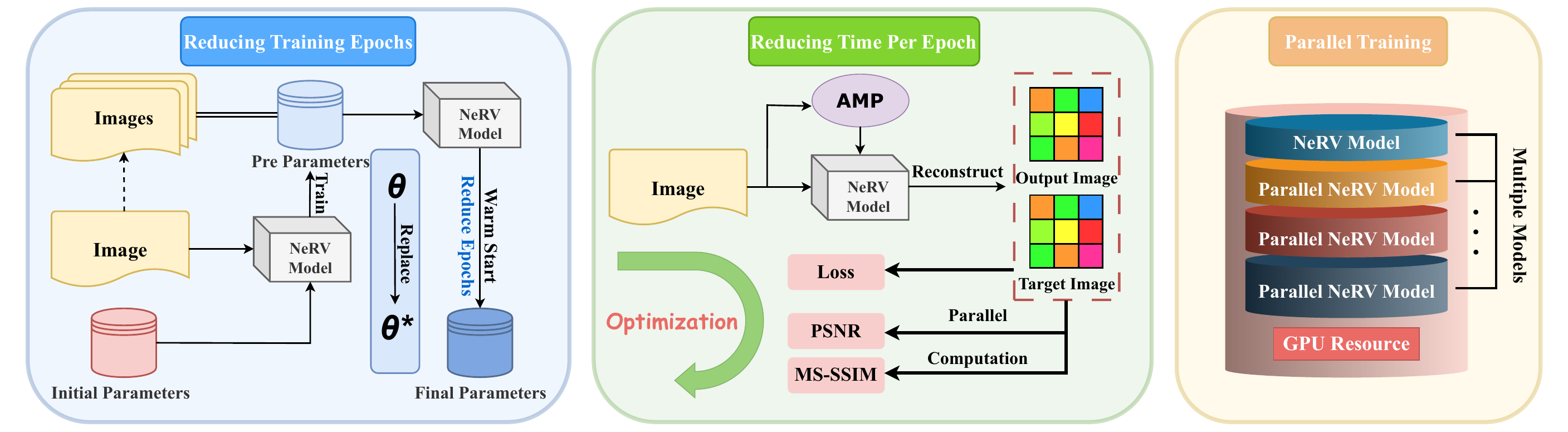} 
        \caption{
Illustration of the proposed three-stage acceleration strategy for NeRV training. 
\textbf{Left}: Reducing training epochs via pretrained model initialization. 
\textbf{Middle}: Reducing time per epoch using AMP, unified metric computation, and batch-wise loss optimization. 
\textbf{Right}: Parallel training of multiple NeRV models to fully utilize GPU resources.
}
        \label{fig:gpu}
    \end{figure*}

b) \underline{Reducing time per training epoch}. Several optimization techniques are integrated: automatic mixed precision (AMP), unified metric computation, and improved loss function design. AMP assigns mixed numerical precisions to operations, performing matrix multiplications and convolutions in FP16 while retaining FP32 for weights and gradients to ensure numerical stability. Dynamic loss scaling prevents underflow, and the system automatically adjusts precision based on hardware capabilities.  

In addition to mixed precision training, we further optimize the training pipeline through the unified metric computation and the improved loss function design. In the original implementation, the PSNR and MS-SSIM metrics are computed separately after each forward pass, using manual loops to process each output patch. This results in redundant post-processing operations and unnecessary overhead when dealing with large batches. To address this inefficiency, we redesign the computation pipeline such that PSNR and MS-SSIM are computed jointly within a vectorized batch-wise process, directly operating on patch outputs and corresponding targets without additional for-loops. This unified metric computation reduces computational cost and accelerates performance monitoring, especially for large patch training.

Moreover, the original loss function relies on the multi-stage interpolation and manual iteration over individual patches, which was inefficient for high-resolution training. In the optimized implementation, we reformulate the loss as a batch-wise patch loss that fully leverages GPU parallelism, eliminating redundant interpolation and looping. Furthermore, we introduce automatic loss weighting across different network stages
to balance optimization dynamics and improve convergence. 
    
c) \underline{Parallel Training of Multiple Models}. The third acceleration strategy takes advantage of the high memory capacity of modern GPUs, enabling multiple NeRV networks to be trained simultaneously on the same device to further improve training efficiency. Although the speedup from parallel training is not strictly linear (due to GPU core contention and memory bandwidth sharing), when sufficient memory is available, multiple models can run in parallel on a single GPU, effectively increasing overall throughput. In practical production environments, especially for batch-processing multiple images, parallel training of multiple models can significantly reduce total training time.

\subsection{Integrate INR with Hyper-Compression}

\begin{figure}[htbp]
        \centering
        \includegraphics[width=0.45\textwidth]{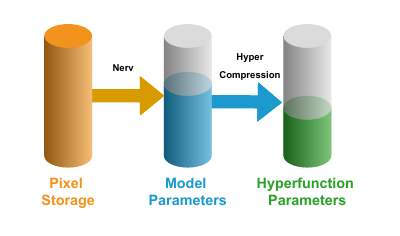} 
        \caption{NeRV encodes images as neural network parameters, which enables to lend post-training model compression techniques to image compresion.}
        \label{fig:compression-hierarchy}
    \end{figure}

Since NeRV encodes images as neural network parameters, it naturally lends itself to model compression techniques. By applying  post-training compression to the NeRV model, we can further reduce storage costs without the need for retraining or additional data preparation. Post-training model compression is lossy, which is acceptable since a good model compression algorithm can already compress a model reasonably well without much distortion of the model's output. Specifically, as Figure \ref{fig:compression-hierarchy} shows, we adopt the Hyper-Compression strategy to pursue a parsimonious representation of the NeRV's parameters.
    
\subsubsection{\textbf{Idea of Hyper-Compression}} The idea of Hyper-Compression originates from the ergodic theory, a branch of dynamical systems, investigates conditions under which the trajectory of a low-dimensional system can densely fill a high-dimensional measure space. This idea suggests that a sufficiently well-behaved transformation $T$ applied repeatedly to a single point $x_0$ can generate a trajectory that approximates any point in a high-dimensional space. Mathematically, the dense trajectory induced by $T$ means that for any $\epsilon$ and $N \in \mathbb{Z} $, there exists $k$ such that
    \begin{equation}
    \Vert \x - T^k(\x_0)\Vert < \epsilon, \quad \x \in \mathbb{R}^N,
    \label{eqn:ergodic_theory}
    \end{equation}
where $\x$ is the high-dimensional target vector, and $k$ is the composition number. 
    
From an engineering standpoint, Eq. \eqref{eqn:ergodic_theory} enables the use of a low-dimensional dynamical system to encode a complex high-dimensional vector. Under the right choice of $T$, we can turn a high-dimensional vector $\x \in \mathbb{R}^N$ into a single scalar $k$, which is a compression if the memory footprint of $k$ is smaller than that of $\x$. Furthermore, if $T$ is a continuous transformation, Eq. \eqref{eqn:ergodic_theory} still holds true. For example, the famous example of irrational winding~\cite{katok1995introduction} shows that
    \begin{theorem}[Katok et al., 1995]
    Let $a_1, \ldots, a_N$ be irrationally independent real numbers. Then for any vector $\{w_n\}_{n=1}^N \subset [0,1]$ and any $\epsilon > 0$, there exists a scalar $\theta^* \in [0, \infty)$ such that:
    \begin{equation}
    \left| w_n - \tau(\theta^* a_n) \right| < \epsilon, \quad \text{for all } n = 1, \ldots, N,
    \label{key}
    \end{equation}
    where $\tau(z) = z - \lfloor z \rfloor$.
    \end{theorem}
    
Hyper-Compression uses the above theorem to design the model compression algorithm. We adopt this Hyper-Compression method and apply it to the INR network.
    
\subsubsection{\textbf{Practical Implementation}}

First, model parameters (e.g., convolutional or fully-connected layer weights) are flattened into one-dimensional vectors and partitioned into fixed-size groups. Then, each group is independently compressed by searching for an optimal scalar $\theta^{*}$ that minimizes the reconstruction error based on Eq.~\eqref{key}, where $a_n=1/(\pi+n), n=1,\cdots,N$. This choice is used to construct the winding trajectory: since $\pi$ is irrational, $\{a_n\}$ are irrationally independent, which avoids periodic repetition and enables a denser and more uniform coverage of the target space. In the Hyper-Compression module, $N$ denotes the dimensionality of each target group; we instantiate the theorem with $N=2$ by pairing adjacent parameters for a 2-to-1 compression, which already provides a favorable compression--fidelity trade-off. The discretization (step size) for quantizing $\theta^{*}$ is selected per experiment to best cover the parameter distribution: a finer step gives denser candidates and smaller $\epsilon$, while a coarser step increases $\epsilon$ but improves compression; for a fixed discretization density, $\epsilon$ typically grows with larger $N$. Subsequently $\theta^{*}$ is quantized into discrete levels with the chosen step size, and only the corresponding index is stored.

Since real-world model parameters are not naturally confined to the interval $[0,1]$, we apply an adaptive affine normalization to map each parameter group into an appropriate target range prior to compression. The normalization range is dynamically selected based on the statistical distribution of the group, which helps balance the compression ratio and reconstruction fidelity. Additionally, for parameter groups exhibiting outliers or heavily-tailed distributions, we incorporate a scaling factor strategy to handle such irregularities. Groups are categorized based on their distance from the group centroid, and appropriate scaling factors are applied to prevent outliers from degrading the overall compression performance.
    
Finally, the decompression process is fully vectorized, which can be parallelized. Each stored $\theta^*$ index is decoded via Eq. \eqref{key}, combined with learned normalization parameters and scaling factors, to accurately reconstruct the corresponding parameter group. This design allows the method to be seamlessly integrated into our compression hierarchy and efficiently applied to large models.
    
Compared to traditional model compression techniques, the Hyper-Compression method has two main merits: First, it can compress a model at a high-compression ratio without the need of re-training, which avoids the catastrophic forgetting and the introducing of the bias into the NeRV model. Second, the compression process takes the shorter time compared to pruning and quantization. For the pruning and quantization, they often need to solve an optimization problem or the re-training, which is time-consuming.

\subsubsection{\textbf{Relation with Vector Quantization}} VQ is a fundamental technique for signal compression. Unlike scalar quantization, which encodes individual values independently, VQ operates on vectors in a high-dimensional space, enabling the joint encoding of multiple correlated features. In the context of learning-based image compression, VQ provides an efficient mechanism to discretize continuous latent representations, thereby facilitating entropy coding and improving compression performance.

Formally, let \(\mathcal{C} = \{\mathbf{e}_1, \mathbf{e}_2, \dots, \mathbf{e}_K\} \subset \mathbb{R}^d\) denote a codebook of \(K\) learnable codewords, where each \(\mathbf{e}_k\) is a \(d\)-dimensional embedding vector. Given an input latent vector \(\mathbf{z} \in \mathbb{R}^d\), the vector quantization process maps \(\mathbf{z}\) to the nearest codeword in the codebook:
\begin{equation}
\text{VQ}(\mathbf{z}) = \mathbf{e}_{k^*}, \quad \text{where} \quad k^* = \arg\min_{k} \|\mathbf{z} - \mathbf{e}_k\|_2.
\label{VQ}
\end{equation}

This nearest-neighbor mapping replaces each latent vector with a discrete codeword, producing a quantized latent representation \(\hat{\mathbf{z}} = \text{VQ}(\mathbf{z})\) based on Eq. \eqref{VQ}. Despite its effectiveness, VQ comes with certain challenges. A large codebook is often required to model diverse content, which increases memory usage. Additionally, during training, some codewords may become unused (a problem known as codebook collapse), leading to inefficient representation capacity. Techniques such as exponential moving average (EMA) updates or commitment loss are commonly used to mitigate these issues.

Unlike VQ, which relies on clustering into a codebook and suffers from scalability issues on large models, Hyper-Compression exploits deterministic trajectories derived from ergodic theory to approximate the parameter space. This eliminates the need for codebook learning and offers better scalability for models with millions of parameters. Moreover, it can avoid querying the codebook in recovering model parameters by directly applying Eq. \eqref{key}.

In summary, as Figure \ref{fig:compression-hierarchy} shows, initially, a large image is stored as pixels, accounting for significant memory. After the compression using NeRV, the input image is transformed into neural network parameters, significantly reducing memory usage while capturing essential image features. Then, Hyper-Compression further compresses the neural network's latent representations, converting them into a few compact hyperparameters, leading to a drastic reduction in the storage size.

\section{Experiments}
To validate the effectiveness of our proposed COLI architecture, we conduct a series of experiments. First, we perform analysis experiments to verify the effectiveness of the core components and design choices of COLI. Second, we conduct comparative experiments against the classical and state-of-the-art image compression methods on standard benchmarks to demonstrate the superior performance of our approach.

\subsection{Analysis Experiments}

In this section, we show that NeRV is an effective backbone for image compression by comparing it with its counterpart method. We also verify that the Hyper-Compression method is compatible with NeRV, \textit{i.e.}, it can further boost the NeRV's compression ratio by nearly 4$\times$, only with minor image quality degradation. 

\subsubsection{\textbf{Experimental Datasets and Models}} 

We conduct image compression experiments and analysis on two models—NeRV~\cite{chen2021nerv} and SIREN~\cite{sitzmann2020implicit}—using a subset of images from the DIV2K dataset~\cite{Agustsson_2017_CVPR_Workshops}. Specifically, we randomly select 10 images with a fixed resolution of 2,040$\times$1,356 and use the same dataset for both models.

NeRV belongs to a coordinate-to-frame INR family, where a convolutional decoder synthesizes image content from compact implicit codes. This design is typically more parameter-efficient for large-image fitting, and it enables fast reconstruction by avoiding per-pixel coordinate regression. In contrast, SIREN is a representative coordinate-to-pixel INR that predicts pixel values individually with a fully-connected MLP. Its pixel-wise formulation is more fine-grained, and can better capture sharp local structures. However, this per-pixel regression mechanism usually demands more parameters and a larger memory footprint for high-resolution images, leading to higher training cost and increased optimization difficulty. This choice allows us to verify the generalization of our compression framework across different INR types: NeRV represents the coordinate-to-frame paradigm, while SIREN covers the per-pixel coordinate regression approach.

\begin{figure*}[htbp]
    \centering
    \begin{subfigure}[b]{0.115\textwidth}
        \includegraphics[width=\linewidth]{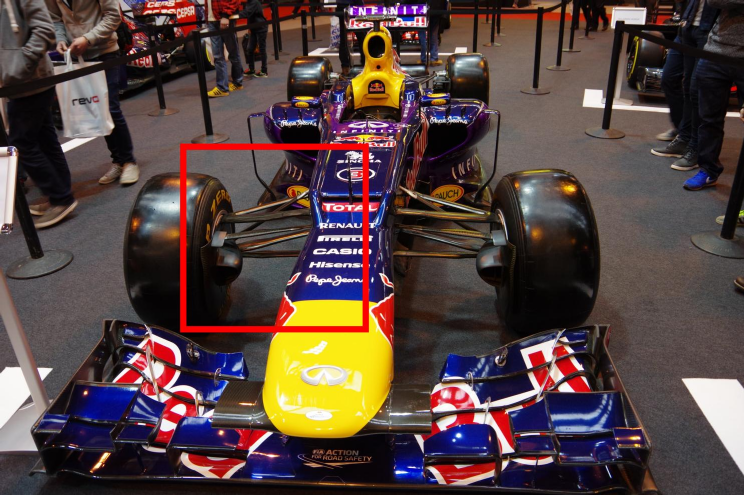}
        \caption{Original}
    \end{subfigure}
    \begin{subfigure}[b]{0.115\textwidth}
        \includegraphics[width=\linewidth]{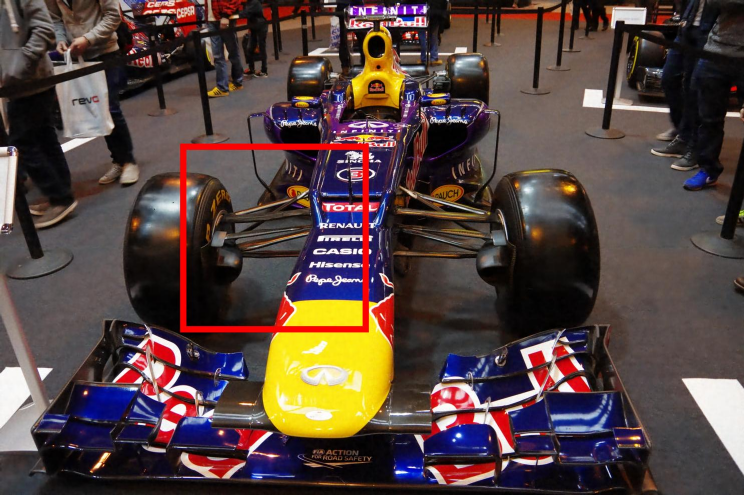}
        \caption{NeRV}
    \end{subfigure}
    \begin{subfigure}[b]{0.115\textwidth}
        \includegraphics[width=\linewidth]{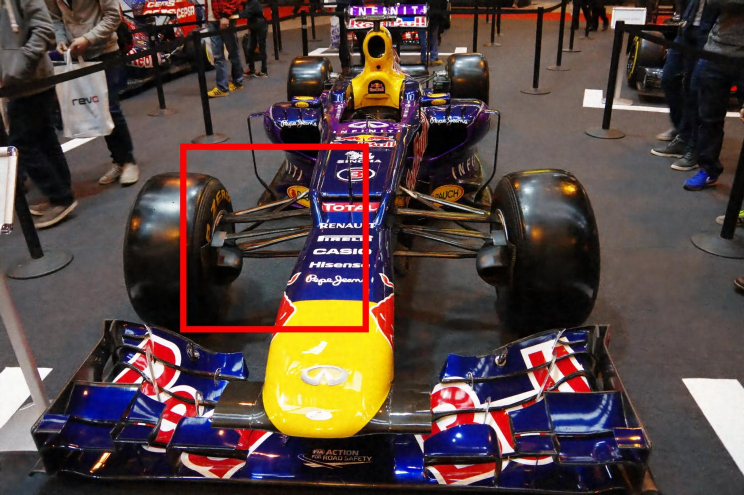}
        \caption{+HC}
    \end{subfigure}
    \begin{subfigure}[b]{0.115\textwidth}
        \includegraphics[width=\linewidth]{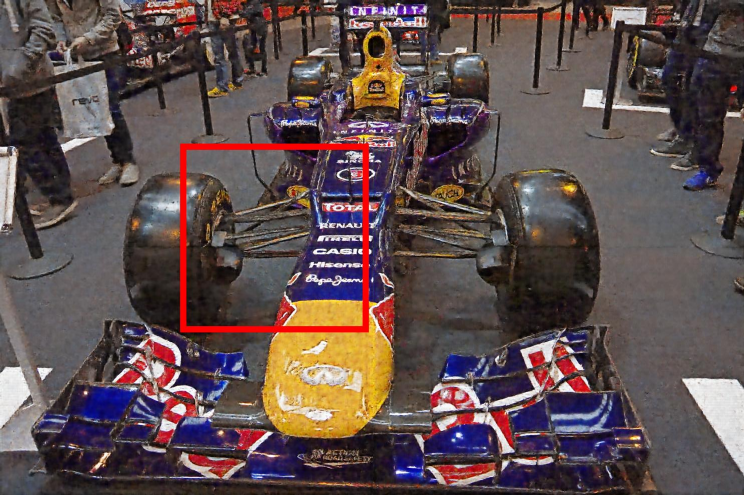}
        \caption{+LR}
    \end{subfigure}
    \begin{subfigure}[b]{0.115\textwidth}
        \includegraphics[width=\linewidth]{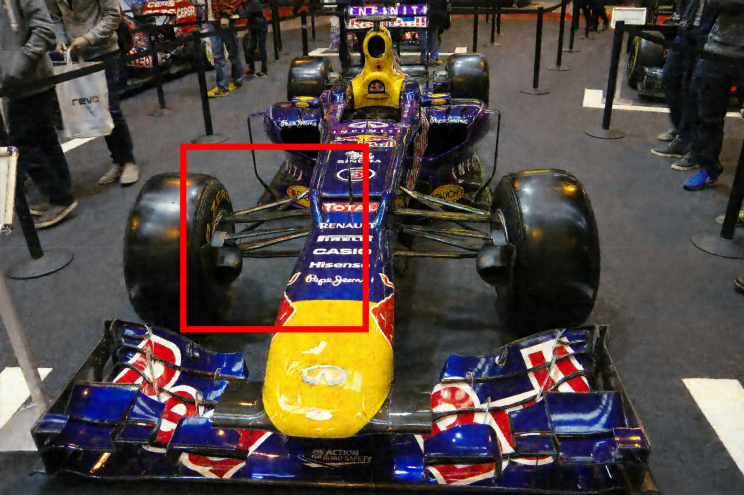}
        \caption{+P}
    \end{subfigure}
    \begin{subfigure}[b]{0.115\textwidth}
        \includegraphics[width=\linewidth]{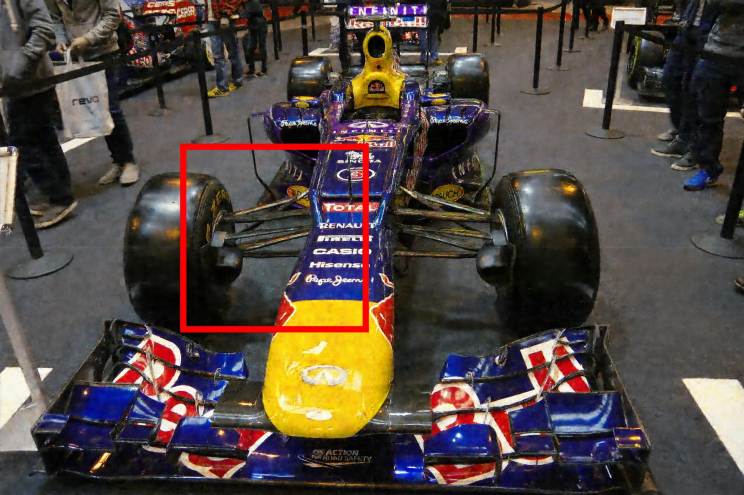}
        \caption{+P+HC}
    \end{subfigure}
    \begin{subfigure}[b]{0.115\textwidth}
        \includegraphics[width=\linewidth]{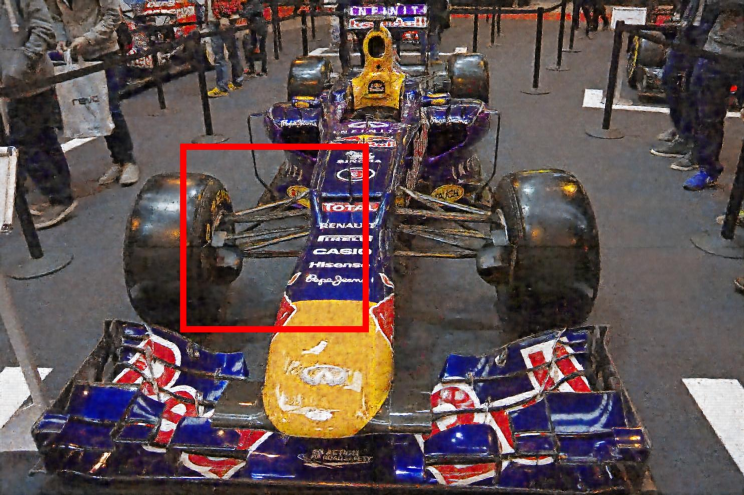}
        \caption{+LR+HC}
    \end{subfigure}
    \begin{subfigure}[b]{0.115\textwidth}
        \includegraphics[width=\linewidth]{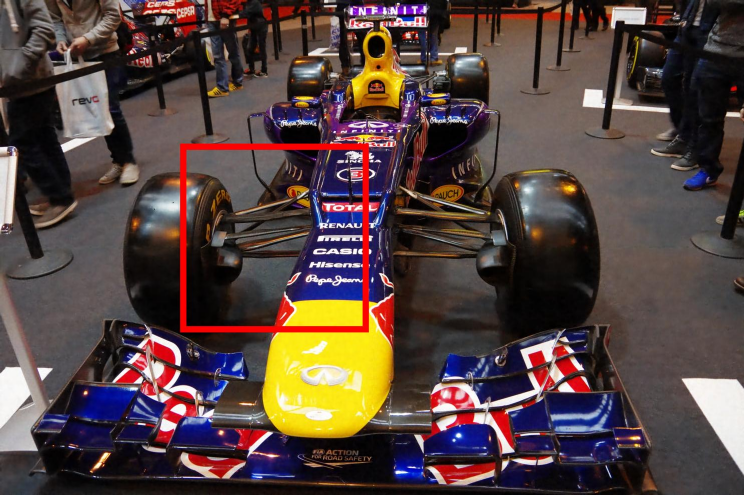}
        \caption{+Q}
    \end{subfigure}
    \caption{A comparison of image reconstruction performance of the NeRV model under different compression strategies.}
    \label{fig:nerv_compression_results}
\end{figure*}

\begin{figure*}[htbp]
    \centering
    \begin{subfigure}[b]{0.115\textwidth}
        \includegraphics[width=\linewidth]{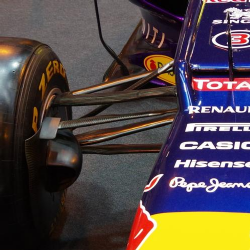}
        \caption{Original}
    \end{subfigure}
    \begin{subfigure}[b]{0.115\textwidth}
        \includegraphics[width=\linewidth]{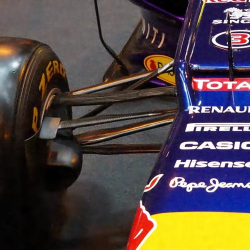}
        \caption{NeRV}
    \end{subfigure}
    \begin{subfigure}[b]{0.115\textwidth}
        \includegraphics[width=\linewidth]{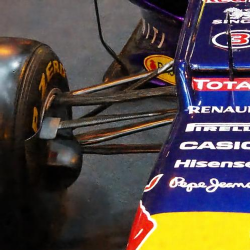}
        \caption{+HC}
    \end{subfigure}
    \begin{subfigure}[b]{0.115\textwidth}
        \includegraphics[width=\linewidth]{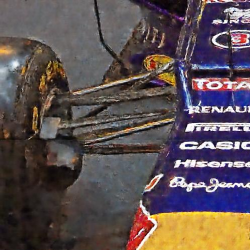}
        \caption{+LR}
    \end{subfigure}
    \begin{subfigure}[b]{0.115\textwidth}
        \includegraphics[width=\linewidth]{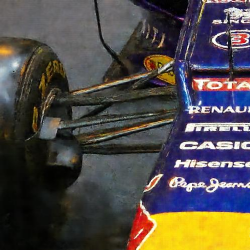}
        \caption{+P}
    \end{subfigure}
    \begin{subfigure}[b]{0.115\textwidth}
        \includegraphics[width=\linewidth]{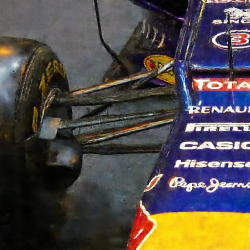}
        \caption{+P+HC}
    \end{subfigure}
    \begin{subfigure}[b]{0.115\textwidth}
        \includegraphics[width=\linewidth]{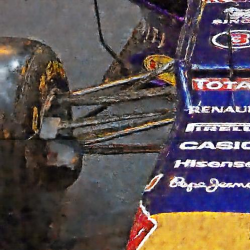}
        \caption{+LR+HC}
    \end{subfigure}
    \begin{subfigure}[b]{0.115\textwidth}
        \includegraphics[width=\linewidth]{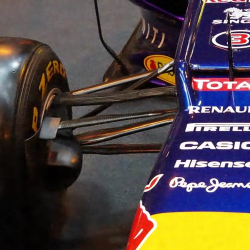}
        \caption{+Q}
    \end{subfigure}
    \caption{Zoomed-in view of the highlighted region in Figure~\ref{fig:nerv_compression_results}.}
    \label{fig:nerv_zoom}
    \vspace{-0.2cm}
\end{figure*}

\subsubsection{\textbf{Compression Strategies}} 
After finishing the training of NeRV and SIREN, we apply various post-training compression methods to their weights without any additional fine-tuning. All results are reported using bit-per-pixel (bpp) as the primary measure of compression efficiency, alongside PSNR, SSIM, and MS-SSIM for reconstruction quality. We compare Hyper-Compression with the following techniques:

\begin{itemize}
  \item \textbf{Pruning~\cite{han2015learning}}:  
  Unstructured pruning based on the $L_1$ norm removes weights with small absolute values. Specifically, we iteratively apply $L_1$ unstructured pruning to all convolutional and fully connected layers. For NeRV, we prune 30\% of weights per layer, while for SIREN we prune 5\% due to its higher sensitivity to sparsity. After pruning, the zeroed weights are kept in the model structure rather than physically removed to maintain compatibility with standard weight loading. We then re-evaluate the sparsity of each layer and save the pruned model for deployment. Since unstructured pruning preserves the dense architecture, the reported bpp is computed from the effective non-zero parameter count implied by the sparsity. This process allows significant parameter reduction while preserving the model architecture, making it suitable for parameter-level compression.

  \item \textbf{Quantization~\cite{jacob2018}}: 
  Post-Training Quantization (PTQ) is employed without retraining. For NeRV, we apply INT8 dynamic quantization to all fully connected layers, converting weights and activations from FP32 to 8-bit integers during inference. For SIREN, we use FP16 precision to reduce storage and inference cost; this setting corresponds to the COIN codec in our context. The quantized models are saved in their entirety, and we compare the resultant file sizes against the original models.

  \item \textbf{Low-rank decomposition~\cite{denton2014exploiting}}: 
  Weight matrices are decomposed via Singular Value Decomposition (SVD), retaining only dominant singular values to approximate the original weights with a lower-rank form. Specifically, we apply SVD to each convolutional and fully connected layer by reshaping their weight tensors into 2D matrices, then keep a predefined number of top singular values, with the effective rank chosen according to the model setting. The resulting low-rank approximation is written back to the original layer weights and evaluated directly, without any retraining. The corresponding bpp is computed from the effective parameter count implied by the selected rank and matrix sizes.
\end{itemize}

\renewcommand{\arraystretch}{1.3} 

We also explore the combinations such as “pruning + Hyper-Compression” and “low-rank decomposition + Hyper-Compression” to evaluate the possibility of stacked model compression to further boost the compression ratio.

\subsubsection{\textbf{NeRV vs SIREN}}

Table~\ref{tab:compression-results} indicates that NeRV shows stronger compression potential and higher reconstruction fidelity on the selected DIV2K images than SIREN. Without compression, NeRV achieves 3.99~bpp with 36.79~dB PSNR and 0.9234 SSIM, while SIREN requires 5.85~bpp and yields a lower PSNR of 32.07~dB. NeRV also remains more robust under post-training compression. For instance, PTQ reduces NeRV’s PSNR only marginally from 36.79~dB to 36.55~dB with the bpp decreasing to 2.29, whereas SIREN under FP16 quantization drops from 32.07~dB to 31.57~dB with a bpp of 3.32. These results suggest that NeRV retains more compressible redundancy in its parameters, while coordinate-wise INRs are more vulnerable to quality loss when their parameters are aggressively compressed.

\subsubsection{\textbf{Efficacy of Different Compression Strategies}} 
We also compare different training-free and data-free post-training compression techniques. Overall, Hyper-Compression is the most effective approach for high-capacity INR models like NeRV. Applying Hyper-Compression alone reduces NeRV’s bpp from 3.99 to 1.06 with only a moderate PSNR drop (36.79~dB to 34.37~dB) and SSIM staying above 0.90. In contrast, pruning and low-rank decomposition lead to substantial fidelity degradation for both models: NeRV with pruning drops to 29.13~dB, while low-rank decomposition further reduces PSNR to 22.05~dB; SIREN is even more sensitive, with pruning and low-rank decomposition producing PSNR around 19~dB; notably, low-rank decomposition results in an effective bpp higher than the uncompressed SIREN. As for quantization, INT8 PTQ reduces NeRV’s bpp to 2.29, and FP16 quantization brings SIREN’s bpp to 3.32; in both cases, the compression gain remains smaller than that achieved by Hyper-Compression. As shown in Figures~\ref{fig:nerv_compression_results} and \ref{fig:nerv_zoom}, the visual comparison aligns with these quantitative scores.

Though Hyper-Compression is a good performer, combining pruning or low-rank decomposition with Hyper-Compression does not further reduce the bpp but indeed degrades image quality, suggesting that stacking multiple post-training compression techniques leads to the amplification of errors. How to effectively combine different orthogonal compression strategies is still an open question.

\begin{table}[ht]
\centering
\caption{Compression results for NeRV and SIREN with different methods. HC = Hyper Compression, P = Pruning, LR = Low Rank Decomposition, Q = Post-Training Quantization.}
\small
\small
\setlength{\tabcolsep}{3.8pt}
\renewcommand{\arraystretch}{1.1}

\begin{tabular}{lccccc}
\hline
\textbf{Model} & \textbf{Method} & \textbf{PSNR} & \textbf{SSIM} & \textbf{MS-SSIM} & \textbf{Bpp} \\
\hline
\multirow{7}{*}{NeRV}
 & \cellcolor{bg01} NeRV~\cite{chen2021nerv} & \cellcolor{bg01} 36.79 & \cellcolor{bg01} 0.9234 & \cellcolor{bg01} 0.9785 & \cellcolor{bg01} 3.99 \\
 & \cellcolor{bg01} + HC~\cite{fan2024hyper} & \cellcolor{bg01} 34.37 & \cellcolor{bg01} 0.9099 & \cellcolor{bg01} 0.9718 & \cellcolor{bg01} 1.06 \\
 & \cellcolor{bg01} + P~\cite{han2015learning} & \cellcolor{bg01} 29.13 & \cellcolor{bg01} 0.8471 & \cellcolor{bg01} 0.9378 & \cellcolor{bg01} 2.79 \\
 & \cellcolor{bg01} + LR~\cite{denton2014exploiting} & \cellcolor{bg01} 22.05 & \cellcolor{bg01} 0.7248 & \cellcolor{bg01} 0.8705 & \cellcolor{bg01} 3.22 \\
 & \cellcolor{bg03} + P + HC~\cite{fan2024hyper} & \cellcolor{bg03} 28.46 & \cellcolor{bg03} 0.8372 & \cellcolor{bg03} 0.9308 & \cellcolor{bg03} 1.06 \\
 & \cellcolor{bg03} + LR + HC~\cite{fan2024hyper} & \cellcolor{bg03} 21.77 & \cellcolor{bg03} 0.7167 & \cellcolor{bg03} 0.8644 & \cellcolor{bg03} 1.06 \\
 & \cellcolor{bg01} + Q~\cite{jacob2018} & \cellcolor{bg01} 36.55 & \cellcolor{bg01} 0.9225 & \cellcolor{bg01} 0.9778 & \cellcolor{bg01} 2.29 \\
\hline
\multirow{7}{*}{SIREN}
 & \cellcolor{bg02} SIREN~\cite{sitzmann2020implicit} & \cellcolor{bg02} 32.07 & \cellcolor{bg02} 0.8698 & \cellcolor{bg02} 0.9601 & \cellcolor{bg02} 5.85 \\
 & \cellcolor{bg02} + HC~\cite{fan2024hyper} & \cellcolor{bg02} 29.96 & \cellcolor{bg02} 0.8290 & \cellcolor{bg02} 0.9024 & \cellcolor{bg02} 2.00 \\
 & \cellcolor{bg02} + P~\cite{han2015learning} & \cellcolor{bg02} 18.58 & \cellcolor{bg02} 0.4940 & \cellcolor{bg02} 0.4505 & \cellcolor{bg02} 5.57 \\
 & \cellcolor{bg02} + LR~\cite{denton2014exploiting} & \cellcolor{bg02} 19.81 & \cellcolor{bg02} 0.5526 & \cellcolor{bg02} 0.5024 & \cellcolor{bg02} 10.47 \\
 & \cellcolor{bg03} + P + HC~\cite{fan2024hyper} & \cellcolor{bg03} 18.52 & \cellcolor{bg03} 0.4906 & \cellcolor{bg03} 0.4461 & \cellcolor{bg03} 2.00 \\
 & \cellcolor{bg03} + LR + HC~\cite{fan2024hyper} & \cellcolor{bg03} 19.70 & \cellcolor{bg03} 0.5474 & \cellcolor{bg03} 0.4965 & \cellcolor{bg03} 2.00 \\
 & \cellcolor{bg02} + Q~\cite{jacob2018} & \cellcolor{bg02} 31.57 & \cellcolor{bg02} 0.8495 & \cellcolor{bg02} 0.9541 & \cellcolor{bg02} 3.32 \\
\hline
\end{tabular}
\vspace{-0.2cm}
\label{tab:compression-results}
\end{table}

\subsection{Comparative Experiments}

\subsubsection{\textbf{Experimental Datasets and Models}} 

In this section, we compare COLI with other image compression methods, including traditional methods, learning-based methods, and the state-of-the-art INR methods, on two representative datasets. The first dataset is the CIL dataset~\cite{howard2012cil}, which consists of high-resolution scanning electron micrographs provided by Louisa Howard, featuring detailed morphological structures of various species such as \textit{Rosa rugosa} and \textit{Helianthus annuus}. Each image is in TIFF format, and we randomly selected images with a resolution of 3,232~$\times$~2,570 pixels for our experiments. For learning-based methods like Balle18 and Cheng20, which require specific input size multiples to fit their encoder architectures, we padded the original CIL images to 3264~$\times$~2624 (Balle18) and 3328~$\times$~2688 (Cheng20), respectively, and restored the reconstructions to the original image resolution after compression.

In addition, we use the CT Heart Segmentation dataset~\cite{NikhilTomar}, which contains a series of 2D computed tomography (CT) heart scans with the resolution of 512~$\times$~512. This dataset provides clear anatomical structures of the human heart, and is widely used in medical image segmentation and compression tasks. Its inherent slice-based nature naturally aligns with our patch-based INR compression strategy, allowing us to further evaluate the effectiveness of COLI for medical images with rich structural information.
\begin{figure*}[htbp]
    \centering
    \begin{subfigure}[b]{0.115\textwidth}
        \includegraphics[width=\linewidth]{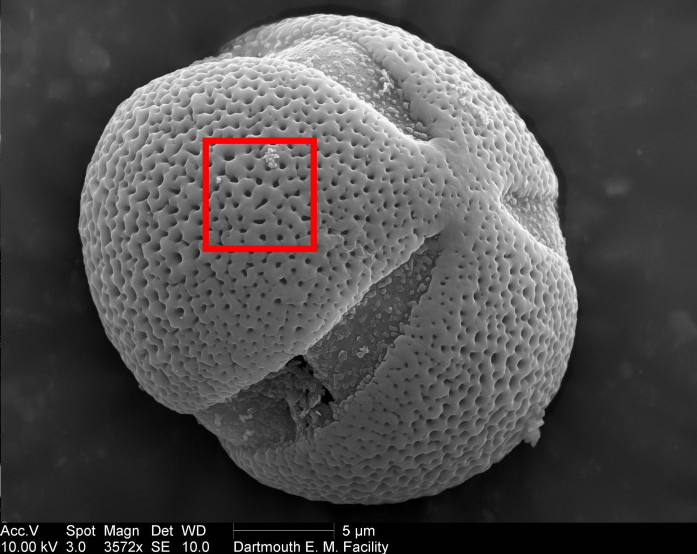}
        \caption{JPEG}
    \end{subfigure}
    \begin{subfigure}[b]{0.115\textwidth}
        \includegraphics[width=\linewidth]{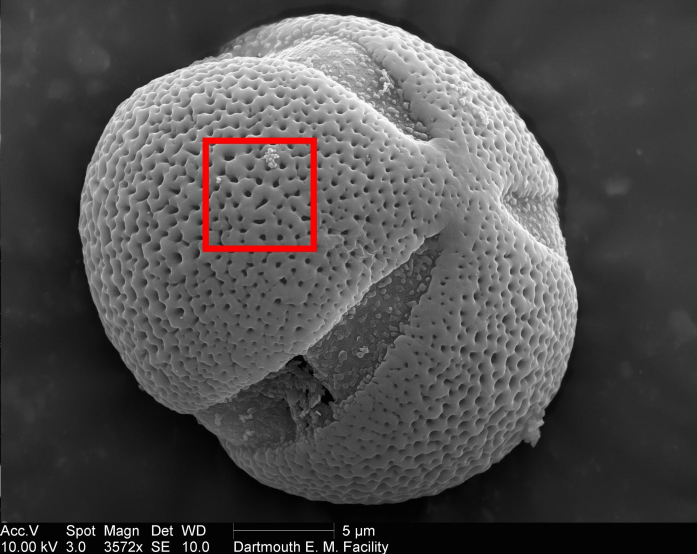}
        \caption{AGPic}
    \end{subfigure}
    \begin{subfigure}[b]{0.115\textwidth}
        \includegraphics[width=\linewidth]{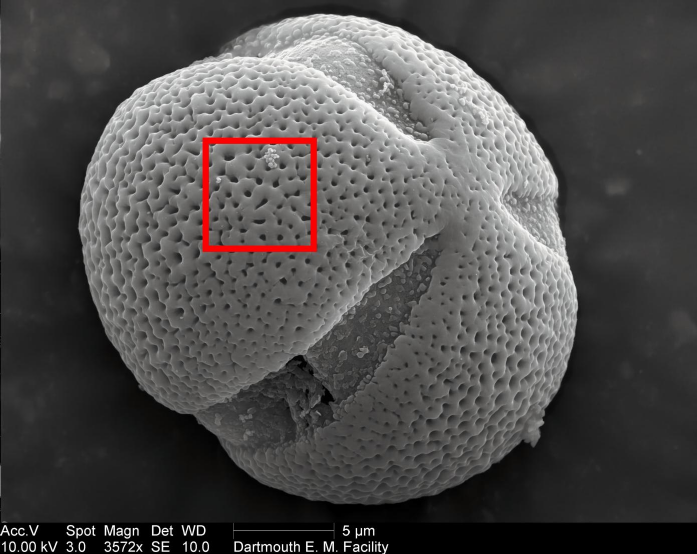}
        \caption{Balle18}
    \end{subfigure}
    \begin{subfigure}[b]{0.115\textwidth}
        \includegraphics[width=\linewidth]{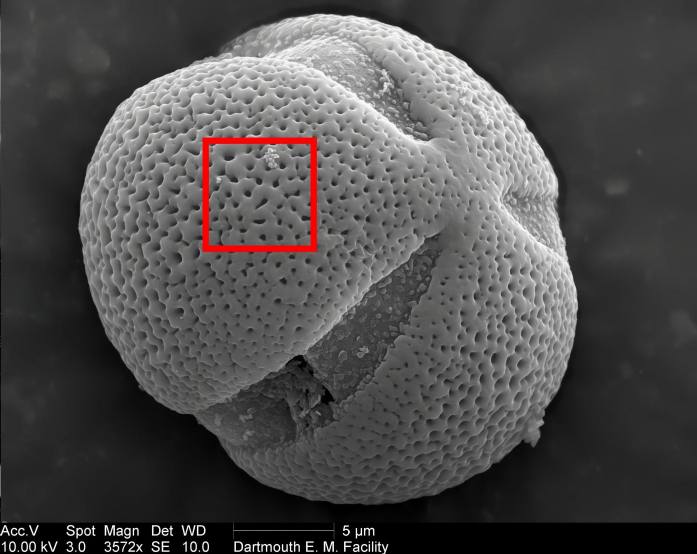}
        \caption{Cheng20}
    \end{subfigure}
    \begin{subfigure}[b]{0.115\textwidth}
        \includegraphics[width=\linewidth]{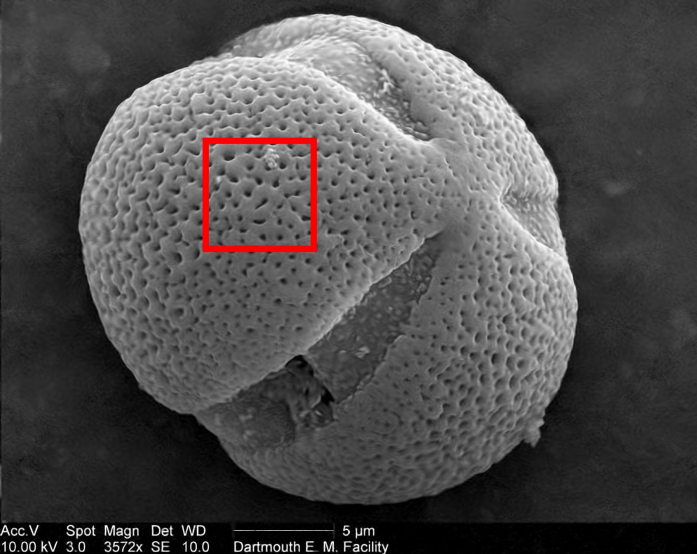}
        \caption{COIN}
    \end{subfigure}
    \begin{subfigure}[b]{0.115\textwidth}
        \includegraphics[width=\linewidth]{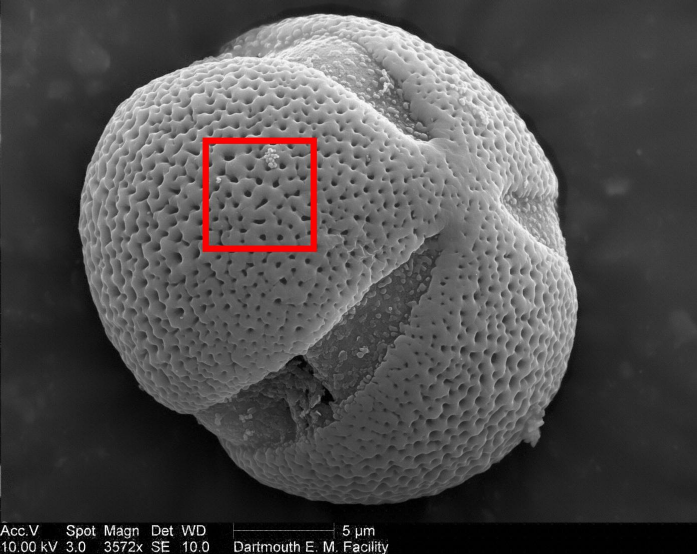}
        \caption{COIN++}
    \end{subfigure}
    \begin{subfigure}[b]{0.115\textwidth}
        \includegraphics[width=\linewidth]{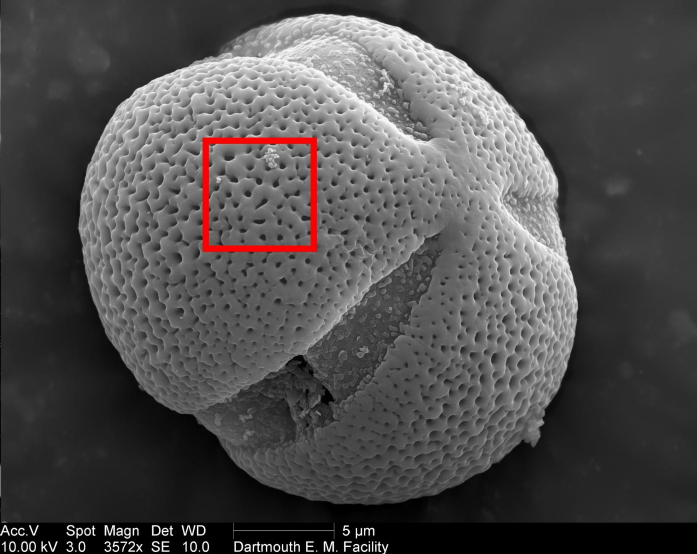}
        \caption{NeRV}
    \end{subfigure}
    \begin{subfigure}[b]{0.115\textwidth}
        \includegraphics[width=\linewidth]{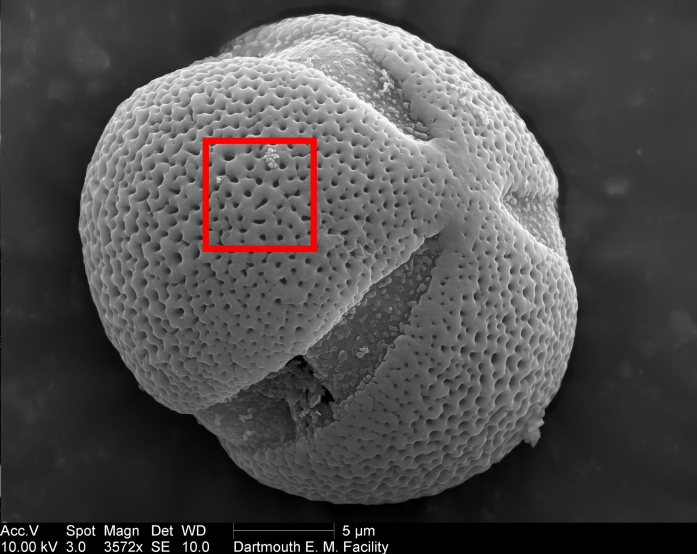}
        \caption{COLI}
    \end{subfigure}
    \caption{The visual comparison of different compression methods on CIL image ID: 40290.}
    \label{fig:cmp_40290}
\end{figure*}

\begin{figure*}[htbp]
    \centering
    \begin{subfigure}[b]{0.115\textwidth}
        \includegraphics[width=\linewidth]{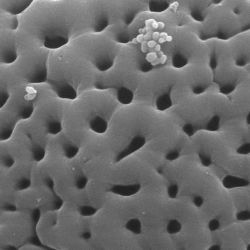}
        \caption{JPEG}
    \end{subfigure}
    \begin{subfigure}[b]{0.115\textwidth}
        \includegraphics[width=\linewidth]{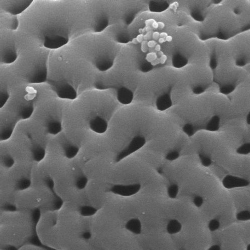}
        \caption{AGPic}
    \end{subfigure}
    \begin{subfigure}[b]{0.115\textwidth}
        \includegraphics[width=\linewidth]{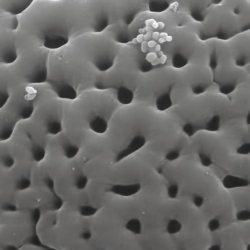}
        \caption{Balle18}
    \end{subfigure}
    \begin{subfigure}[b]{0.115\textwidth}
        \includegraphics[width=\linewidth]{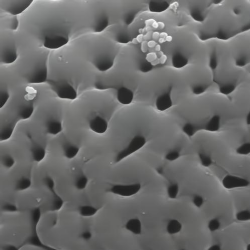}
        \caption{Cheng20}
    \end{subfigure}
    \begin{subfigure}[b]{0.115\textwidth}
        \includegraphics[width=\linewidth]{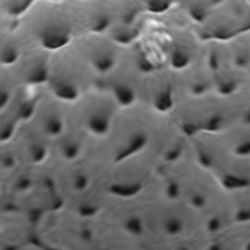}
        \caption{COIN}
    \end{subfigure}
    \begin{subfigure}[b]{0.115\textwidth}
        \includegraphics[width=\linewidth]{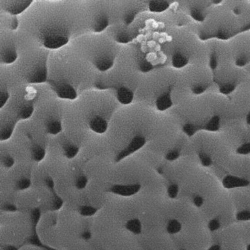}
        \caption{COIN++}
    \end{subfigure}
    \begin{subfigure}[b]{0.115\textwidth}
        \includegraphics[width=\linewidth]{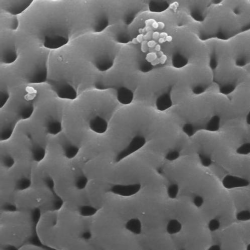}
        \caption{NeRV}
    \end{subfigure}
    \begin{subfigure}[b]{0.115\textwidth}
        \includegraphics[width=\linewidth]{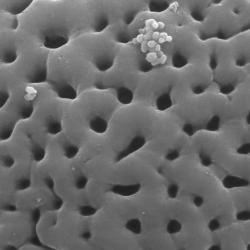}
        \caption{COLI}
    \end{subfigure}
    \caption{Zoomed-in view of the highlighted region in Figure~\ref{fig:cmp_40290}.}
    \label{fig:cmp_40290_}
    \vspace{-0.2cm}
\end{figure*}

\begin{table}[ht]
\centering
\caption{Performance comparison of different compression methods on the CIL dataset\label{tab:results_6}.}
\small
\setlength{\tabcolsep}{3.1pt}  
\renewcommand{\arraystretch}{1.05} 
\begin{tabular}{lccccccc}
\toprule
\textbf{Models} & \multicolumn{2}{c}{\textbf{PSNR}} & \multicolumn{2}{c}{\textbf{SSIM}} & \multicolumn{2}{c}{\textbf{MS-SSIM}} & \textbf{Bpp} \\
& Mean & Std & Mean & Std & Mean & Std & \\
\midrule
\cellcolor{bg01}JPEG~\cite{wallace1992jpeg} & \cellcolor{bg01}37.60 & \cellcolor{bg01}0.94 & \cellcolor{bg01}0.8985 & \cellcolor{bg01}0.0231 & \cellcolor{bg01}0.9885 & \cellcolor{bg01}0.0024 & \cellcolor{bg01}0.67 \\
\cellcolor{bg01}AGPic~\cite{AGPicCompress2025} & \cellcolor{bg01}38.22 & \cellcolor{bg01}0.55 & \cellcolor{bg01}0.9264 & \cellcolor{bg01}0.0129 & \cellcolor{bg01}0.9929 & \cellcolor{bg01}0.0010 & \cellcolor{bg01}1.88 \\
\cellcolor{bg02}Balle18~\cite{balle2018variational} & \cellcolor{bg02}36.04 & \cellcolor{bg02}0.87 & \cellcolor{bg02}0.8769 & \cellcolor{bg02}0.0374 & \cellcolor{bg02}0.9799 & \cellcolor{bg02}0.0060 & \cellcolor{bg02}0.35 \\
\cellcolor{bg02}Cheng20~\cite{cheng} & \cellcolor{bg02}36.43 & \cellcolor{bg02}1.00 & \cellcolor{bg02}0.8762 & \cellcolor{bg02}0.0411 & \cellcolor{bg02}0.9784 & \cellcolor{bg02}0.0074 & \cellcolor{bg02}0.24 \\
\cellcolor{bg03}COIN~\cite{dupont2021coin} & \cellcolor{bg03}27.06 & \cellcolor{bg03}1.13 & \cellcolor{bg03}0.6989 & \cellcolor{bg03}0.0455 & \cellcolor{bg03}0.8722 & \cellcolor{bg03}0.0300 & \cellcolor{bg03}0.41 \\
\cellcolor{bg03}COIN++~\cite{dupont2022coinpp} & \cellcolor{bg03}29.40 & \cellcolor{bg03}0.38 & \cellcolor{bg03}0.7194 & \cellcolor{bg03}0.0214 & \cellcolor{bg03}0.9390 & \cellcolor{bg03}0.0034 & \cellcolor{bg03}1.82 \\
\cellcolor{bg03}NeRV~\cite{chen2021} & \cellcolor{bg03}34.83 & \cellcolor{bg03}0.73 & \cellcolor{bg03}0.8873 & \cellcolor{bg03}0.0342 & \cellcolor{bg03}0.9821 & \cellcolor{bg03}0.0057 & \cellcolor{bg03}2.41 \\
\cellcolor{bg03}\textbf{COLI} & \cellcolor{bg03}\textbf{33.85} & \cellcolor{bg03}\textbf{0.79} & \cellcolor{bg03}\textbf{0.8712} & \cellcolor{bg03}\textbf{0.0344} & \cellcolor{bg03}\textbf{0.9754} & \cellcolor{bg03}\textbf{0.0065} & \cellcolor{bg03}\textbf{0.56} \\
\bottomrule
\end{tabular}
\end{table}

\begin{table}[ht]
\centering
\caption{Performance comparison of different compression methods on the CT Heart Segmentation Dataset dataset\label{tab:results_3}.}
\small
\setlength{\tabcolsep}{3.1pt}  
\renewcommand{\arraystretch}{1.05} 
\begin{tabular}{lccccccc}
\toprule
\textbf{Models} & \multicolumn{2}{c}{\textbf{PSNR}} & \multicolumn{2}{c}{\textbf{SSIM}} & \multicolumn{2}{c}{\textbf{MS-SSIM}} & \textbf{Bpp} \\
& Mean & Std & Mean & Std & Mean & Std & \\
\midrule
\cellcolor{bg01}JPEG~\cite{wallace1992jpeg} & \cellcolor{bg01}39.75 & \cellcolor{bg01}1.04 & \cellcolor{bg01}0.9452 & \cellcolor{bg01}0.0152 & \cellcolor{bg01}0.9945 & \cellcolor{bg01}0.0016 & \cellcolor{bg01}0.85 \\
\cellcolor{bg01}AGPic~\cite{AGPicCompress2025} & \cellcolor{bg01}39.69 & \cellcolor{bg01}0.94 & \cellcolor{bg01}0.9562 & \cellcolor{bg01}0.0240 & \cellcolor{bg01}0.9962 & \cellcolor{bg01}0.0010 & \cellcolor{bg01}1.64 \\
\cellcolor{bg02}Balle18~\cite{balle2018variational} & \cellcolor{bg02}34.63 & \cellcolor{bg02}2.14 & \cellcolor{bg02}0.8370 & \cellcolor{bg02}0.0575 & \cellcolor{bg02}0.9625 & \cellcolor{bg02}0.0648 & \cellcolor{bg02}0.18 \\
\cellcolor{bg02}Cheng20~\cite{cheng} & \cellcolor{bg02}35.23 & \cellcolor{bg02}0.80 & \cellcolor{bg02}0.8643 & \cellcolor{bg02}0.0337 & \cellcolor{bg02}0.9688 & \cellcolor{bg02}0.0069 & \cellcolor{bg02}0.12 \\
\cellcolor{bg03}COIN~\cite{dupont2021coin} & \cellcolor{bg03}31.44 & \cellcolor{bg03}0.66 & \cellcolor{bg03}0.7564 & \cellcolor{bg03}0.0322 & \cellcolor{bg03}0.9384 & \cellcolor{bg03}0.0090 & \cellcolor{bg03}2.17 \\
\cellcolor{bg03}COIN++~\cite{dupont2022coinpp} & \cellcolor{bg03}30.43 & \cellcolor{bg03}0.72 & \cellcolor{bg03}0.6889 & \cellcolor{bg03}0.0284 & \cellcolor{bg03}0.9297 & \cellcolor{bg03}0.0036 & \cellcolor{bg03}1.36 \\
\cellcolor{bg03}NeRV~\cite{chen2021} & \cellcolor{bg03}38.03 & \cellcolor{bg03}0.96 & \cellcolor{bg03}0.9058 & \cellcolor{bg03}0.0289 & \cellcolor{bg03}0.9823 & \cellcolor{bg03}0.0055 & \cellcolor{bg03}2.75 \\
\cellcolor{bg03}\textbf{COLI} & \cellcolor{bg03}\textbf{37.28} & \cellcolor{bg03}\textbf{0.85} & \cellcolor{bg03}\textbf{0.8996} & \cellcolor{bg03}\textbf{0.0285} & \cellcolor{bg03}\textbf{0.9802} & \cellcolor{bg03}\textbf{0.0054} & \cellcolor{bg03}\textbf{0.56} \\
\bottomrule
\end{tabular}
\end{table}

For a fair comparison, we include two traditional codecs: JPEG~\cite{wallace1992jpeg} and AGPic~\cite{AGPicCompress2025}; two learning-based neural codecs: Balle18~\cite{balle2018variational} and Cheng20~\cite{cheng}; and three representative INR-based codecs: COIN~\cite{dupont2021coin}, COIN++~\cite{dupont2022coinpp}, and NeRV~\cite{chen2021}, together with our proposed COLI. COIN++ requires cross-image training. Following its standard setup, we split the images into approximately $2/3$ for training and $1/3$  for testing, and all COIN++ results reported in the tables are measured on the held-out test set. All methods are evaluated using PSNR, SSIM, and MS-SSIM for reconstruction quality, together with bit-per-pixel (bpp) to reflect compression efficiency.

\begin{figure}[htbp]
    \centering
\includegraphics[width=0.5\textwidth]{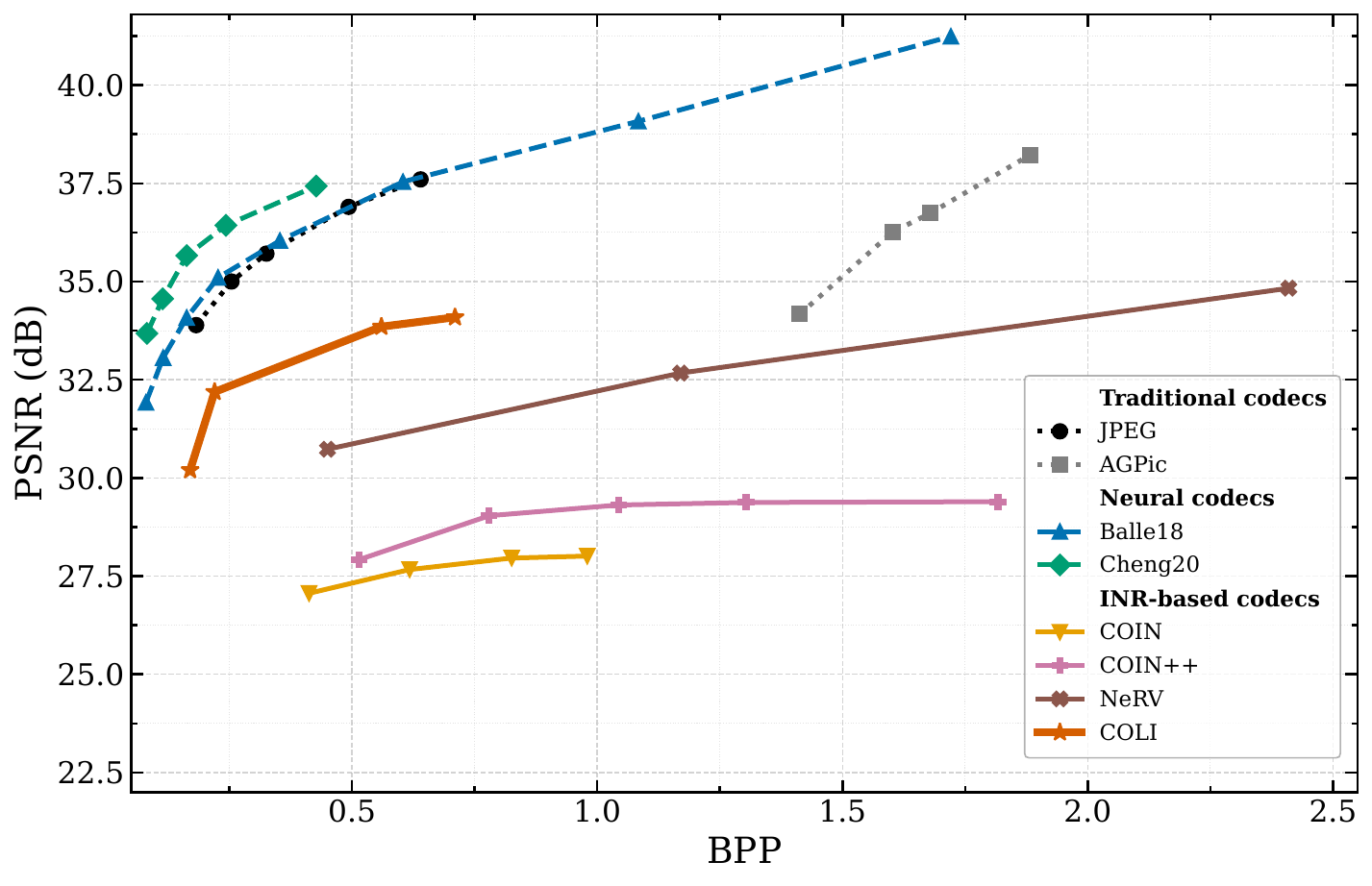}
    \caption{Schematic diagram of the bpp–PSNR curves on the CIL Dataset.}
    \label{fig:bpp_psnr_curve}
    \vspace{-0.4cm}
    \end{figure}

\subsubsection{\textbf{Quantitative Results and Analysis}} 

Figure~\ref{fig:bpp_psnr_curve} plots the overall bpp--PSNR trend on the CIL dataset; detailed numerical comparisons on both datasets are reported in Tables~\ref{tab:results_6} and~\ref{tab:results_3}. Notably, the relative ordering of bpp across CIL and CT is dataset-dependent, which is a natural consequence of the distinct rate--distortion characteristics of these two modalities. Since bpp essentially measures the entropy of the learned representation, it is highly sensitive to signal statistics. The CIL images exhibit substantially richer textures and more pronounced high-frequency content, thereby increasing latent entropy and leading to higher bpp for learning-based codecs. In contrast, CT heart slices are visually smoother and more repetitive, yielding more compact latent distributions that are easier to model and compress, and thus resulting in lower bpp.

\begin{figure*}[htbp]
    \centering
    \begin{subfigure}[b]{0.115\textwidth}
        \includegraphics[width=\linewidth]{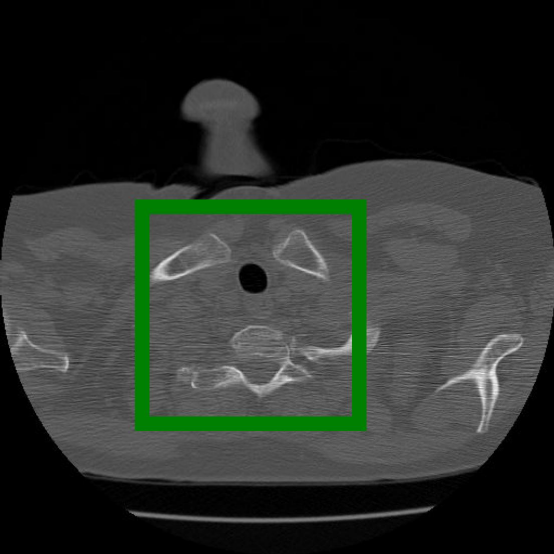}
        \caption{JPEG}
    \end{subfigure}
    \begin{subfigure}[b]{0.115\textwidth}
        \includegraphics[width=\linewidth]{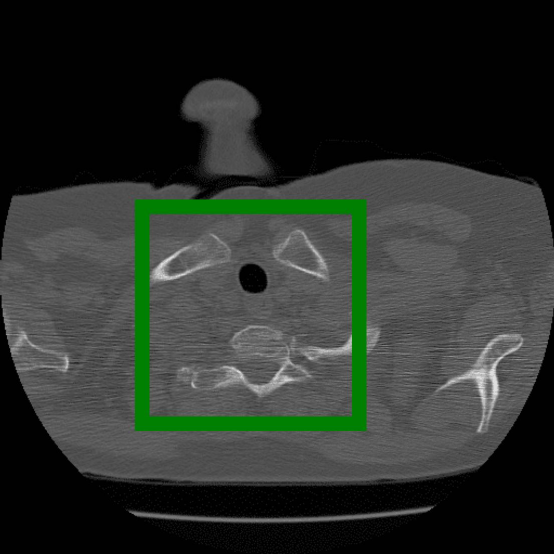}
        \caption{AGPic}
    \end{subfigure}
    \begin{subfigure}[b]{0.115\textwidth}
        \includegraphics[width=\linewidth]{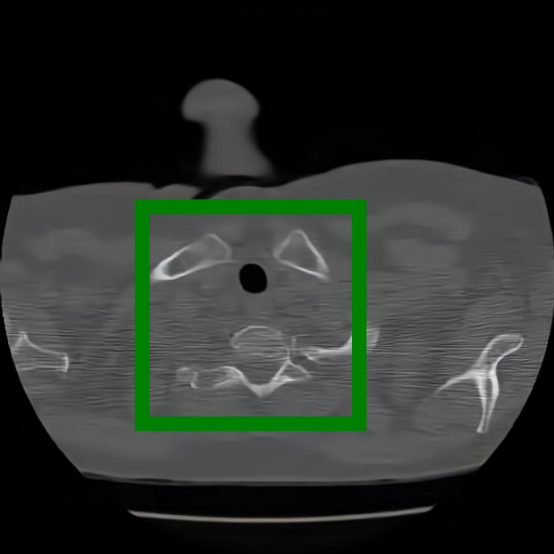}
        \caption{Balle18}
    \end{subfigure}
    \begin{subfigure}[b]{0.115\textwidth}
        \includegraphics[width=\linewidth]{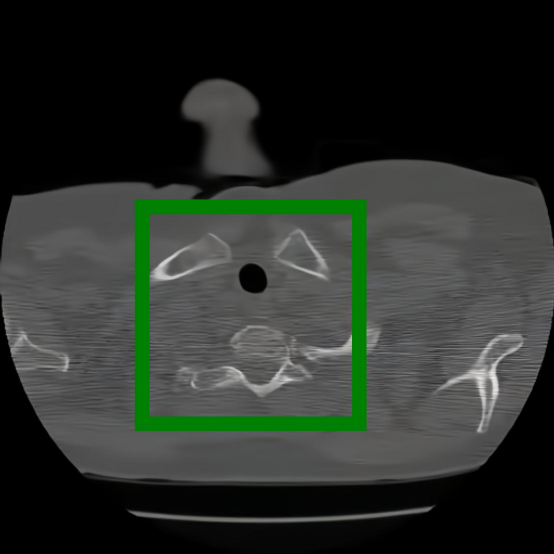}
        \caption{Cheng20}
    \end{subfigure}
    \begin{subfigure}[b]{0.115\textwidth}
        \includegraphics[width=\linewidth]{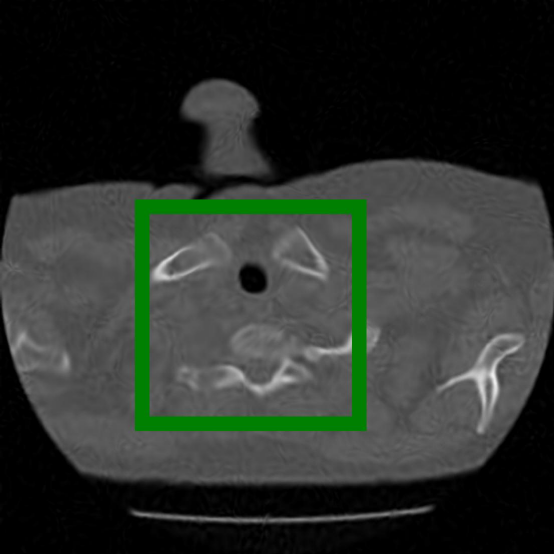}
        \caption{COIN}
    \end{subfigure}
    \begin{subfigure}[b]{0.115\textwidth}
        \includegraphics[width=\linewidth]{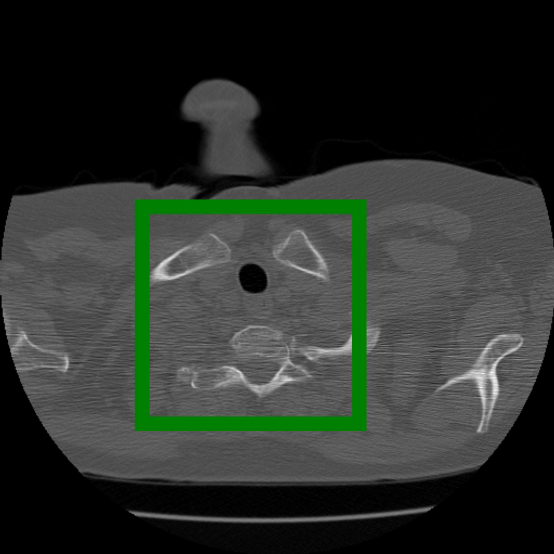}
        \caption{COIN++}
    \end{subfigure}
    \begin{subfigure}[b]{0.115\textwidth}
        \includegraphics[width=\linewidth]{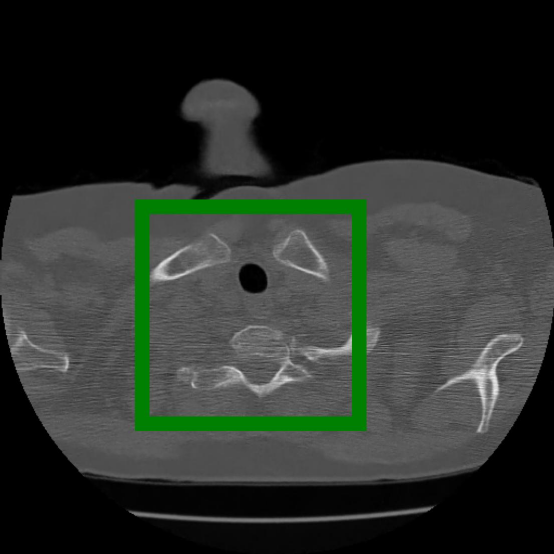}
        \caption{NeRV}
    \end{subfigure}
    \begin{subfigure}[b]{0.115\textwidth}
        \includegraphics[width=\linewidth]{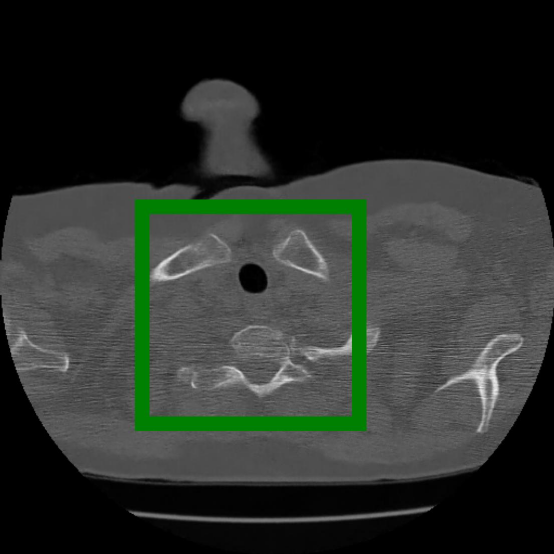}
        \caption{COLI}
    \end{subfigure}
    \caption{The visual comparison of different compression methods on a CT Heart Segmentation image.}
    \label{fig:cmp_ct1}
\end{figure*}

\begin{figure*}[htbp]
    \centering
    \begin{subfigure}[b]{0.115\textwidth}
        \includegraphics[width=\linewidth]{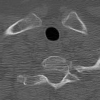}
        \caption{JPEG}
    \end{subfigure}
    \begin{subfigure}[b]{0.115\textwidth}
        \includegraphics[width=\linewidth]{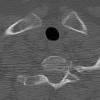}
        \caption{AGPic}
    \end{subfigure}
    \begin{subfigure}[b]{0.115\textwidth}
        \includegraphics[width=\linewidth]{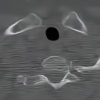}
        \caption{Balle18}
    \end{subfigure}
    \begin{subfigure}[b]{0.115\textwidth}
        \includegraphics[width=\linewidth]{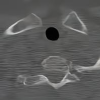}
        \caption{Cheng20}
    \end{subfigure}
    \begin{subfigure}[b]{0.115\textwidth}
        \includegraphics[width=\linewidth]{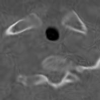}
        \caption{COIN}
    \end{subfigure}
    \begin{subfigure}[b]{0.115\textwidth}
        \includegraphics[width=\linewidth]{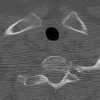}
        \caption{COIN++}
    \end{subfigure}
    \begin{subfigure}[b]{0.115\textwidth}
        \includegraphics[width=\linewidth]{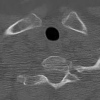}
        \caption{NeRV}
    \end{subfigure}
    \begin{subfigure}[b]{0.115\textwidth}
        \includegraphics[width=\linewidth]{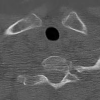}
        \caption{COLI}
    \end{subfigure}
    \caption{Zoomed-in view of the highlighted region in Figure~\ref{fig:cmp_ct1}.}
    \label{fig:cmp_ct1_zoom}
\end{figure*}

\begin{figure*}[htbp]
    \centering
    \begin{subfigure}[b]{0.115\textwidth}
        \includegraphics[width=\linewidth]{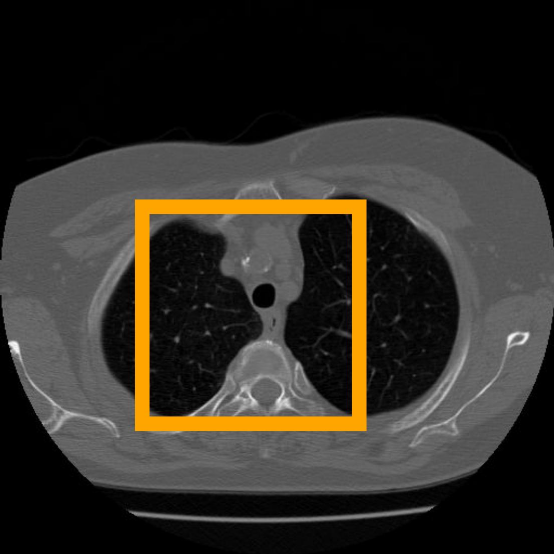}
        \caption{JPEG}
    \end{subfigure}
    \begin{subfigure}[b]{0.115\textwidth}
        \includegraphics[width=\linewidth]{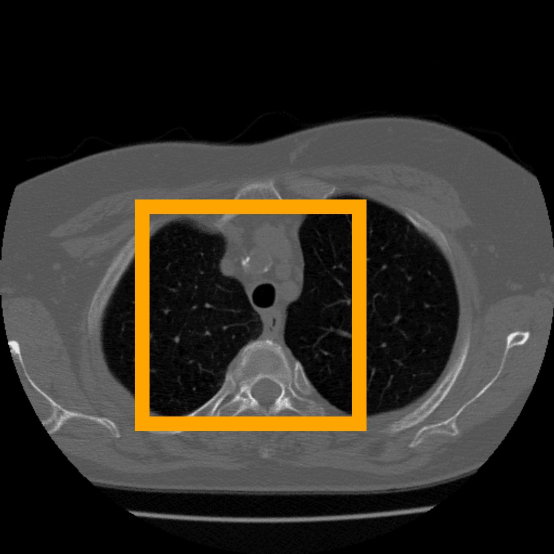}
        \caption{AGPic}
    \end{subfigure}
    \begin{subfigure}[b]{0.115\textwidth}
        \includegraphics[width=\linewidth]{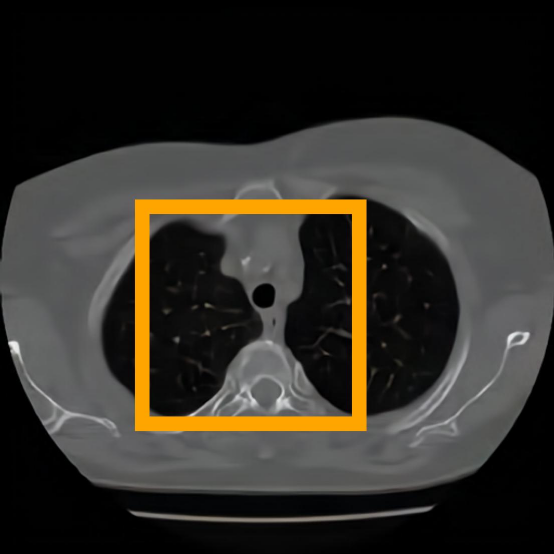}
        \caption{Balle18}
    \end{subfigure}
    \begin{subfigure}[b]{0.115\textwidth}
        \includegraphics[width=\linewidth]{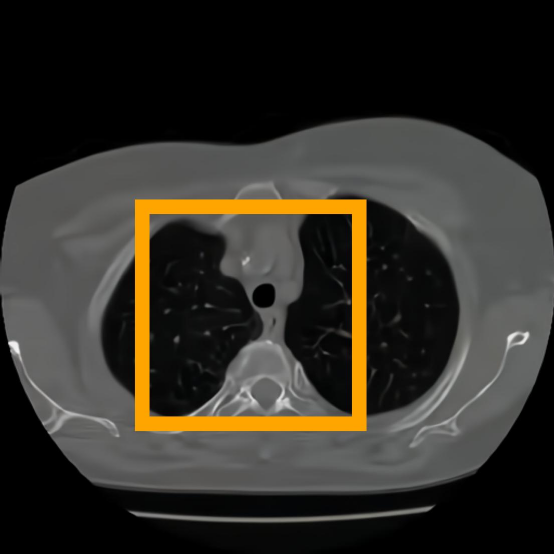}
        \caption{Cheng20}
    \end{subfigure}
    \begin{subfigure}[b]{0.115\textwidth}
        \includegraphics[width=\linewidth]{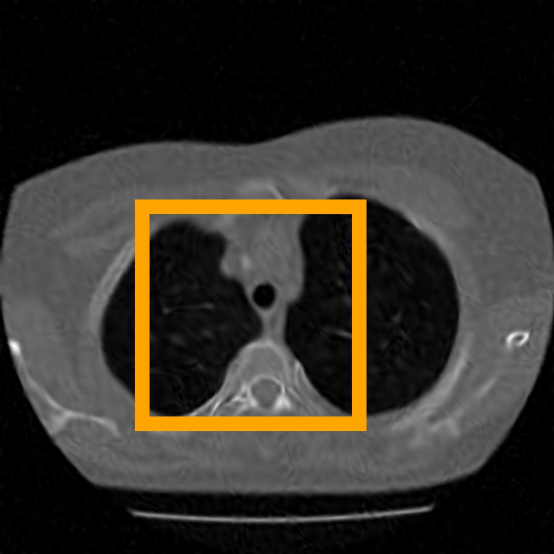}
        \caption{COIN}
    \end{subfigure}
    \begin{subfigure}[b]{0.115\textwidth}
        \includegraphics[width=\linewidth]{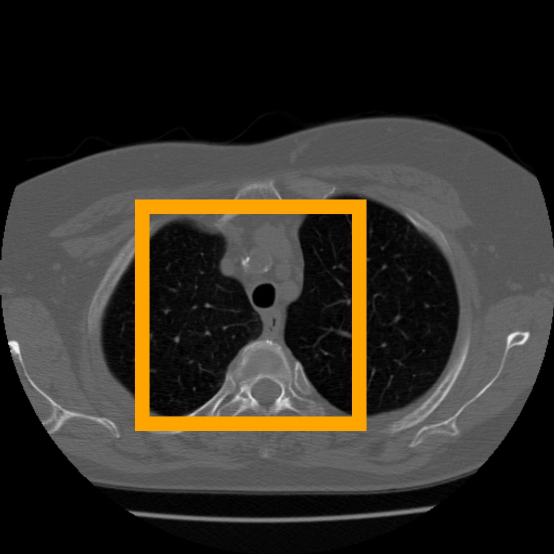}
        \caption{COIN++}
    \end{subfigure}
    \begin{subfigure}[b]{0.115\textwidth}
        \includegraphics[width=\linewidth]{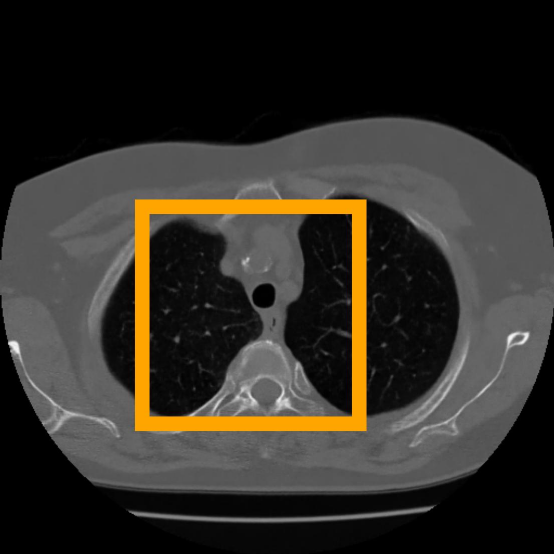}
        \caption{NeRV}
    \end{subfigure}
    \begin{subfigure}[b]{0.115\textwidth}
        \includegraphics[width=\linewidth]{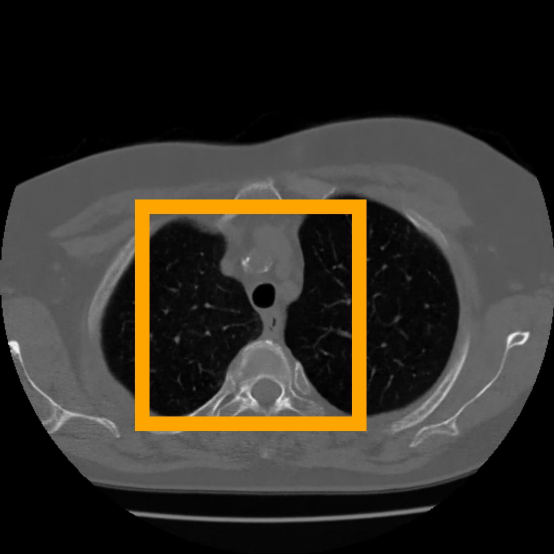}
        \caption{COLI}
    \end{subfigure}
    \caption{The visual comparison of different compression methods on another CT Heart Segmentation image.}
    \label{fig:cmp_ct2}
\end{figure*}

\begin{figure*}[htbp]
    \centering
    \begin{subfigure}[b]{0.115\textwidth}
        \includegraphics[width=\linewidth]{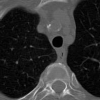}
        \caption{JPEG}
    \end{subfigure}
    \begin{subfigure}[b]{0.115\textwidth}
        \includegraphics[width=\linewidth]{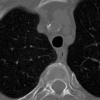}
        \caption{AGPic}
    \end{subfigure}
    \begin{subfigure}[b]{0.115\textwidth}
        \includegraphics[width=\linewidth]{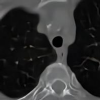}
        \caption{Balle18}
    \end{subfigure}
    \begin{subfigure}[b]{0.115\textwidth}
        \includegraphics[width=\linewidth]{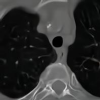}
        \caption{Cheng20}
    \end{subfigure}
    \begin{subfigure}[b]{0.115\textwidth}
        \includegraphics[width=\linewidth]{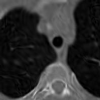}
        \caption{COIN}
    \end{subfigure}
    \begin{subfigure}[b]{0.115\textwidth}
        \includegraphics[width=\linewidth]{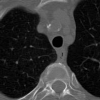}
        \caption{COIN++}
    \end{subfigure}
    \begin{subfigure}[b]{0.115\textwidth}
        \includegraphics[width=\linewidth]{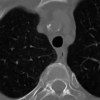}
        \caption{NeRV}
    \end{subfigure}
    \begin{subfigure}[b]{0.115\textwidth}
        \includegraphics[width=\linewidth]{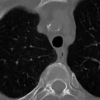}
        \caption{COLI}
    \end{subfigure}
    \caption{Zoomed-in view of the highlighted region in Figure~\ref{fig:cmp_ct2}.}
    \label{fig:cmp_ct2_zoom}
    \vspace{-0.5cm}
\end{figure*}

We begin with traditional codecs. JPEG~\cite{wallace1992jpeg} remains a strong classical baseline, delivering high reconstruction fidelity at its available operating points. AGPic~\cite{AGPicCompress2025} also provides acceptable quality at moderate bitrates, but its handcrafted pipeline is less effective in the low-bpp regime; in our setting it cannot be pushed to low bitrates without a clear loss of detail, and it is consistently outperformed by COLI at comparable or lower bpp. Moreover, JPEG performs competitively within the tested bitrate range. However, its bitrate control is relatively discrete, and pushing to extremely low bpp often leads to unstable visual quality. This limitation is particularly relevant for large-scale archival scenarios where more compression is required.

Among learning-based approaches, Balle18~\cite{balle2018variational} (implemented as the \texttt{bmshj2018\_hyperprior} architecture) and Cheng20~\cite{cheng} (using the \texttt{cheng2020\_anchor} variant) represent typical end-to-end neural compression pipelines. On CIL, Cheng20 reaches high fidelity at low bpp, and similar trends are observed on CT. However, in our experiments, these methods occasionally fail to produce reliable reconstructions for certain images, and their strict input-size constraints increase deployment complexity. As a result, although learning-based codecs are strong in rate--distortion terms, their practical robustness on diverse large images can be less predictable.

We finally conduct comparisons with representative INR-based models, including COIN~\cite{dupont2021coin}, COIN++~\cite{dupont2022coinpp}, NeRV~\cite{chen2021}, and COLI. COIN and COIN++ are coordinate-network baselines, with COIN++ trained on approximately $2/3$ of the images and evaluated on the held-out test split. NeRV serves as a stronger INR baseline but typically operates at a much higher bitrate to preserve quality. Across both datasets, COLI establishes a clear advantage within the INR family: on CIL, COLI achieves 33.85~dB at 0.56~bpp, clearly surpassing COIN and COIN++ in fidelity while using a substantially lower bitrate than NeRV; on CT Heart, COLI maintains high reconstruction quality at 0.56~bpp, whereas COIN and COIN++ require notably higher bitrates to reach much lower PSNR. In brief, these results confirm that COLI advances INR-based compression toward a more practical extreme-low-bpp regime, while retaining stable reconstructions and a lightweight neural code compared with end-to-end neural codecs.

\subsubsection{\textbf{Visual Comparison}}  

Figures~\ref{fig:cmp_40290} and~\ref{fig:cmp_40290_} present representative examples from the CIL dataset and their zoomed-in regions, respectively. For the CT Heart Segmentation dataset, Figures~\ref{fig:cmp_ct1} and~\ref{fig:cmp_ct2} illustrate typical reconstruction results, while Figures~\ref{fig:cmp_ct1_zoom} and~\ref{fig:cmp_ct2_zoom} further enlarge local regions to compare structural details. Across both datasets, COLI preserves fine structures and global continuity well even at relatively low bpp. Specifically, on the CIL dataset, COLI attains the lowest bitrate among INR-based methods while still maintaining clear texture details. On the CT Heart dataset, COLI also operates at a low bpp within the INR family and delivers superior visual quality with improved detail fidelity and smoother structural presentation. The zoomed-in regions show fewer blocking artifacts and smoother transitions, making COLI especially suitable for large images and medical images. These visual results validate that INR-based compression can achieve efficient storage with reliable perceptual consistency, offering practical advantages for real-world large-scale image processing.

\subsubsection{\textbf{Training Acceleration}} To validate the effectiveness of the proposed three-facet training acceleration strategy in COLI, we conduct experiments from three perspectives: pretraining-based transfer, per-epoch acceleration, and memory analysis for parallel multi-model training. To highlight the effectiveness of the proposed acceleration strategy, we report the overall training time savings achieved by COLI.

\begin{figure}[htbp]
    \centering
\includegraphics[width=0.5\textwidth]{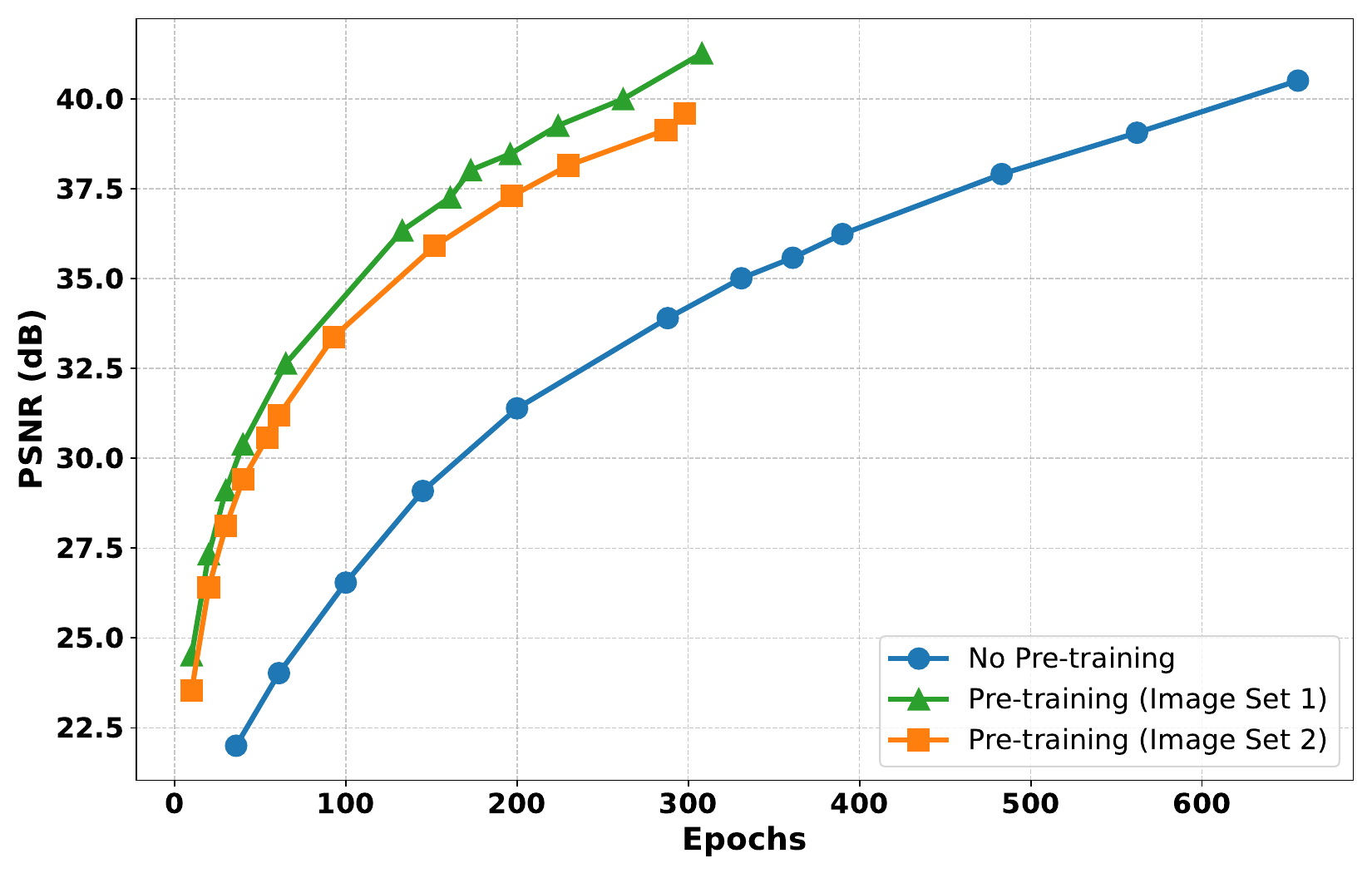}
    \caption{A comparison between the training curves of pretrained models and the same model under random initialization.}
    \label{fig:pre}
    \vspace{-0.2cm}
    \end{figure}

To evaluate the effect of model pretraining on INR-based compression, we conducted an experiment using a collection of CT slices from The Cancer Imaging Archive (TCIA)~\cite{clark2013cancer}. We randomly selected three subsets of CT images, each containing 216 slices, and compressed each subset individually. One subset was trained from scratch without any pretraining, while the other two used NeRV models initialized with pretrained weights from similar CT slices.

Figure~\ref{fig:pre} shows that the pretrained NeRV models converge faster and reach higher PSNR with fewer epochs. For example, the pretrained model (Image Set 1) achieves approximately 39dB PSNR after only 200 epochs, whereas the baseline without pretraining needs over 600 epochs to reach comparable quality. This demonstrates that pretraining is especially effective for medical images with high content consistency, where similar structures across slices can be leveraged to reduce training time and improve convergence stability.

    \begin{table}[ht]
    \centering
    \caption{Epoch-wise acceleration time (Unit: seconds, time measured as average over epochs and shortest epoch)\label{tab:results}.}
    \small
    \setlength{\tabcolsep}{6pt} 
    \renewcommand{\arraystretch}{1} 
    \renewcommand{\cellgape}{\Gape[1.5pt]} 
    \begin{tabular}{lcccc}
    \hline
    \textbf{Model} 
      & \textbf{10-epoch}
      & \textbf{20-epoch}
      & \textbf{30-epoch}
      & \textbf{Shortest} \\
    \hline
    NeRV-S & 32.25 & 31.78 & 31.59 & 30.83 \\
    NeRV-S + ACC & 29.34 & 29.04 & 28.88 & 28.31 \\
    NeRV-M & 32.40 & 31.98 & 31.84 & 31.27 \\
    NeRV-M + ACC & 29.25 & 28.96 & 28.83 & 28.33 \\
    \hline
    \end{tabular}
    \vspace{-0.2cm}
    \end{table}

Table~\ref{tab:results} presents the average training time per epoch for NeRV-S and NeRV-M with and without AMP and unified metric computation. Each experiment uses the same number of images instead of the entire dataset. The results show that, with these optimizations, the average per-epoch training time is reduced by approximately 10\%, which demonstrates a practical improvement.

We analyze the resource usage for parallel multi-model training. When GPU memory is sufficient, training a single NeRV-S network consumes about 1600 MiB, while the accelerated version uses approximately 2700 MiB. By fully utilizing available memory, multiple networks can be trained in parallel on the same GPU, further reducing total training time and improving efficiency for batch processing tasks. 
    \begin{table}[ht]
    \centering
    \caption{Training time and speedup ratio comparison under similar PSNR levels.\label{tab:results_2}.}
    \small
    \setlength{\tabcolsep}{6pt} 
    \renewcommand{\arraystretch}{1} 
    \renewcommand{\cellgape}{\Gape[1.5pt]} 
    \begin{tabular}{lcc}
    \hline
    \textbf{Models} & \textbf{Training (mins)} & \textbf{Speedup ratio} \\
    \hline
    \cellcolor{bg01}ORI (PSNR $\approx$ 30)   & \cellcolor{bg01}98  & \cellcolor{bg01}/     \\
    \cellcolor{bg01}Ours (PSNR $\approx$ 30)  & \cellcolor{bg01}11  & \cellcolor{bg01}\textbf{8.91x} \\
    \cellcolor{bg02}ORI (PSNR $\approx$ 37)   & \cellcolor{bg02}268 & \cellcolor{bg02}/     \\
    \cellcolor{bg02}Ours (PSNR $\approx$ 37)  & \cellcolor{bg02}64  & \cellcolor{bg02}\textbf{4.19x} \\
    \cellcolor{bg03}ORI (PSNR $\approx$ 40)   & \cellcolor{bg03}375 & \cellcolor{bg03}/     \\
    \cellcolor{bg03}Ours (PSNR $\approx$ 40)  & \cellcolor{bg03}95  & \cellcolor{bg03}\textbf{3.95x} \\
    \hline
    \end{tabular}
\end{table}

Table~\ref{tab:results_2} reports the training time and the speedup ratio under three different PSNR levels. The experiments are conducted on the TCIA dataset by selecting two groups of 216 CT slices for training. In our accelerated setting, two networks are trained concurrently on the same GPU, and the reported time is the average training time of both models. Compared to the original pipeline, our method consistently achieves similar PSNR with significantly less training time. For example, at the target PSNR of around 30~dB, the training time drops from 98 minutes to 11 minutes, yielding a speedup of 8.91×. Similar trends hold for higher PSNR levels, with an overall training time reduction of approximately 4× on average. This demonstrates that our acceleration strategy is highly effective for similar medical image datasets.

\section{Conclusion}

This work has presented COLI, a scalable framework for compressing large images using Implicit Neural Representations (INRs). COLI integrates Hyper-Compression—an effective post-training model compression technique—with a comprehensive acceleration strategy. Specifically, COLI encodes large images with a NeRV-based INR backbone, leveraging pre-trained models, per-epoch acceleration, and parallelized batch training to reduce encoding time. The Hyper-Compression module then further compresses the NeRV parameters, efficiently exploiting internal model redundancies. Evaluations on electron microscopy and medical imaging datasets demonstrate that COLI achieves a favorable rate–distortion trade-off in the low-bpp regime, clearly improving over prior INR-based codecs while maintaining stable reconstructions, and providing a lightweight neural code compared with end-to-end neural compressors. Future research will explore extending COLI to 3D volumetric settings to better exploit inter-slice correlations in medical scans, and incorporating lightweight entropy coding for Hyper-Compression indices to further reduce the bitrate.

\section*{Acknowledgments}
All authors declare that they have no known conflicts of interest in terms of competing financial interests or personal relationships that could have an influence or are relevant to the work reported in this paper.


\bibliographystyle{IEEEtran}
\bibliography{reference2}

\end{document}